\documentclass[lettersize,journal]{IEEEtran}
\usepackage{amsmath,amsfonts,amssymb}
\usepackage{algorithmic}
\usepackage{algorithm}
\usepackage{booktabs}
\usepackage{multicol}
\usepackage{multirow}
\usepackage{booktabs}
\usepackage{enumerate}
\usepackage{xcolor}
\usepackage{colortbl}
\usepackage{tabularray}
\usepackage{microtype}
\usepackage[utf8]{inputenc}
\usepackage[T1]{fontenc}
\UseTblrLibrary{booktabs}
\usepackage{tabularx}
\usepackage{hhline}
\usepackage{tikz}
\usepackage{paralist}
\usepackage{bbding}
\usepackage{array}
\usepackage[caption=false,font=scriptsize,labelfont=sf,textfont=sf]{subfig}
\usepackage{textcomp}
\usepackage{stfloats}
\usepackage{soul}
\usepackage{url}
\usepackage{verbatim}
\usepackage{graphicx}
\usepackage{hhline}
\usepackage{cite}
\usepackage[colorlinks,
linkcolor=black,
anchorcolor=black,
citecolor=black]{hyperref}
\usepackage{orcidlink}
\usepackage{amsthm}

\newtheoremstyle{tpamiPhenomenon}  
{\topsep}                        
{\topsep}                        
{}                               
{}                               
{\bfseries}                      
{.}                              
{ }                              
{\thmname{#1}\ \thmnumber{#2}}   

\theoremstyle{tpamiPhenomenon}
\newtheorem{phenomenon}{Phenomenon}

\def\eg{\emph{e.g.}}

\newtheorem{theorem}{\bfseries Theorem}

\begin{document}

\title{Beyond Motion Cues and Structural Sparsity: Revisiting Small Moving Target Detection}

\author{Guoyi~Zhang,~Han~Wang,~and~Xiaohu~Zhang
	\thanks{Guoyi~Zhang, Han~Wang, and Xiaohu~Zhang are with the School of Aeronautics and Astronautics, Sun Yat-sen University, Shenzhen 518107, Guangdong, China.(Corresponding authors: \emph{Han Wang and Xiaohu Zhang}, email:zhanggy57@mail2.sysu.edu.cn;zhangxiaohu@mail.sysu.edu.cn)}
}

\markboth{Journal of \LaTeX\ Class Files,~Vol.~14, No.~8, August~2021}%
{Guoyi Zhang \MakeLowercase{\textit{et al.}}: Beyond Motion Cues and Structural Sparsity: Revisiting Small Moving Target Detection}


\maketitle

\begin{abstract}	
Small moving target detection is crucial for many defense applications but remains highly challenging due to low signal-to-noise ratios, ambiguous visual cues, and cluttered backgrounds. In this work, we propose a novel deep learning framework that differs fundamentally from existing approaches, which often rely on target-specific features or motion cues and lack robustness in complex environments.
Our key insight is that small target detection and background discrimination are inherently coupled: even cluttered video backgrounds exhibit strong low-rank structures that can serve as stable priors for detection.
We reformulate the task as tensor-based low-rank and sparse decomposition, guided by a physically grounded analysis of background, target, and noise. Building on this, we introduce TenRPCANet, a deep network that relies on mild, physically grounded assumptions (local smoothness from the optical point spread function), in contrast to conventional methods that impose task-specific constraints such as Gaussian curvature. 
To realize this framework, we propose a tokenization strategy that implicitly enforces multi-order tensor low-rank priors through a self-attention mechanism. This mechanism captures both local and non-local self-similarity to model the low-rank background without explicit iterative optimization. Inspired by the sparse component update in tensor RPCA, we further design a feature refinement module to enhance target saliency.
The proposed method achieves state-of-the-art performance on two highly distinct and challenging tasks: multi-frame infrared small target detection and space debris detection (visible-light), demonstrating both the effectiveness and generalizability of our approach.
\end{abstract}


\section{Introduction} \label{Section:Introduction}
\IEEEPARstart{S}{mall} moving target detection \cite{11474969,8734113,10914515} is crucial for national defense tasks such as missile early warning \cite{11373245,gao2013infrared,liu2023infrared}, reconnaissance \cite{10946108,UIUNet,DNANet}, and space surveillance \cite{11036127,lin2021new}. However, detecting small targets is challenging due to low signal-to-noise ratios (SNR), severe foreground-background imbalance, diverse target appearances, and lack of distinctive features \cite{9217948,liu2023infrared,CSRNet,11080263}.

Recent deep learning methods often rely on hand-crafted structural priors such as Gaussian curvature \cite{10586900}, local contrast patterns \cite{Dai_2021_WACV,ALCNet}, or target shape \cite{ISNet}, and explicitly exploit motion cues like directional consistency or trajectory smoothness \cite{li2023direction}, achieving promising results on specific benchmarks \cite{11130659, MSHNet}. However, these assumptions are inherently task-specific: Gaussian curvature assumes a fixed target intensity profile; contrast-based methods fail when target-background difference is low; shape-based priors become unreliable when targets and false alarms are highly similar in appearance and false alarms are abundant; and motion-based approaches struggle when targets move irregularly or are too dim to track. As a result, their robustness and generalization are limited \cite{8106808, li2023direction}, leading to poor cross-task transferability \cite{chen2024convolutional}.

\begin{figure}
	\centering
	\includegraphics[width=1.0\linewidth]{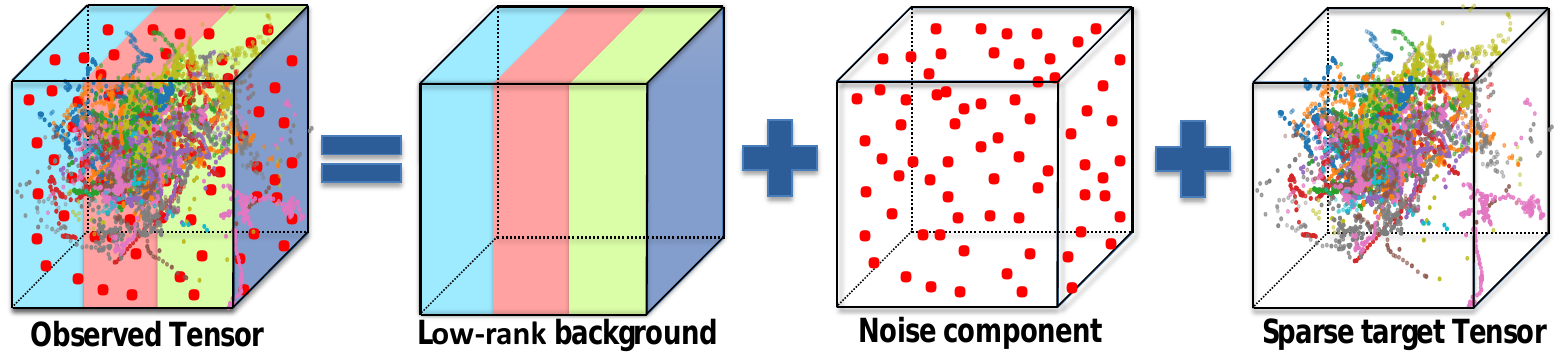}
	\caption{Illustration of the problem formulation. The colored trajectories are visualized from real image sequences \cite{RDIAN} to show that motion patterns are not fixed, indicating that motion cues are inherently non-robust.}
	\label{fig:fig_trpcaill}
\end{figure}

To address the aforementioned challenges, this paper proposes a novel deep learning paradigm for small moving target detection. We observe that target detection and background discrimination are inherently coupled. Due to structural redundancy in video sequences, backgrounds exhibit stable low-rank properties \cite{7488247,6820779,9749010,8017459}, while targets appear as spatiotemporally structured outliers \cite{8543221,6781644,6216381}. Based on this observation, the detection problem can be reformulated as a low-rank and sparse decomposition task (Fig.~\ref{fig:fig_trpcaill}):
\begin{equation}
	\begin{aligned}
		\min_{\boldsymbol{\mathcal{L}},\boldsymbol{\mathcal{S}},\boldsymbol{\mathcal{N}}}\ rank(\boldsymbol{\mathcal{L}})+&\lambda J_{\boldsymbol{\mathcal{S}}}(\boldsymbol{\mathcal{S}}) + \eta J_{\boldsymbol{\mathcal{N}}}(\boldsymbol{\mathcal{N}})\\ \mathrm{s.t.}\quad \boldsymbol{\mathcal{X}}=&\boldsymbol{\mathcal{L}}+\boldsymbol{\mathcal{S}} + \boldsymbol{\mathcal{N}}
		\label{Eq:forma}
	\end{aligned}
\end{equation}
where $\boldsymbol{\mathcal{X}}$ denotes the observed tensor, $\boldsymbol{\mathcal{L}}$ $\boldsymbol{\mathcal{S}}$ and $\boldsymbol{\mathcal{N}}$ correspond to the low-rank background, sparse target components and noise component, respectively. The parameters $\lambda > 0$ and $\eta > 0$ are scalar regularization weights that control the trade-off between the three objective terms. The function $rank(\cdot)$ imposes the low-rank constraint on the background component, while $J_{\boldsymbol{\mathcal{S}}}(\cdot)$ denotes the structured sparsity-inducing regularization applied to the target component. The third term, $J_{\boldsymbol{\mathcal{N}}}(\cdot)$, penalizes the noise component to ensure robustness against background clutter, sensor noise, or modeling errors.

\begin{figure*}[!t]
	\centering
	\includegraphics[width=1\linewidth]{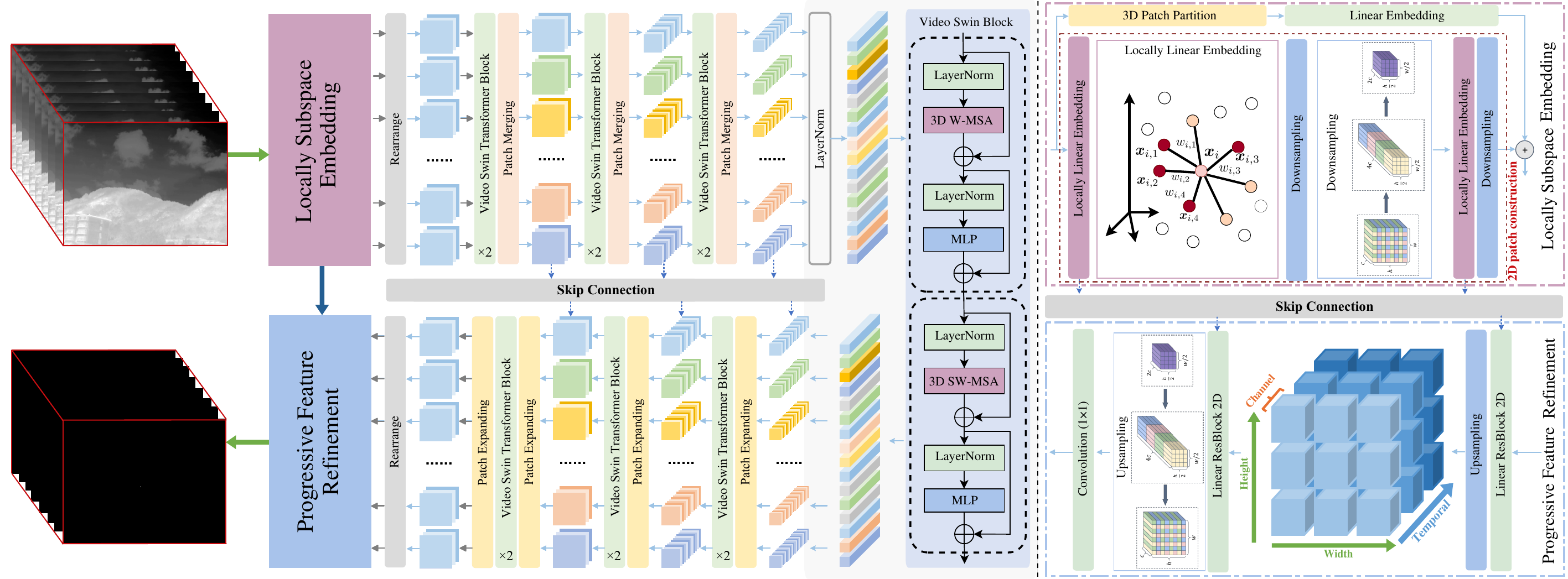}
	\caption{Overall architecture of TenRPCANet. The framework is built upon our physically grounded analysis and consists of three core components.
		First, the Locally Subspace Embedding (LSE) serves as a geometry-aware tokenization module. It embeds input spatiotemporal patches into a latent manifold, preserving local neighborhood structures while implicitly encoding third- and fourth-order tensor low-rank priors. This embedding shapes the underlying manifold geometry and regularizes the subsequent self-attention mechanism without modifying its parametric form.
		Second, the encoder and decoder, constructed from Video Swin Transformer blocks \cite{liu2022video}, operate on the embedded tokens. Due to the spatial downsampling in LSE (stride 4), small target signals are largely suppressed, allowing the self-attention to focus on capturing the low-rank structure of the background via spatiotemporal self-similarity.
		Third, the Progressive Feature Refinement (PFR) module, inspired by the sparse component update in tensor RPCA, refines target features using lightweight operations, enforcing structural sparsity guided by the local intensity smoothness prior (due to the point spread function).}
	\label{fig:OverallofArch}
\end{figure*}

Next, we conduct a physically grounded analysis of the background, foreground, and noise components in Eq.~(\ref{Eq:forma}), relying on mild and physically grounded assumptions about the target (PSF-induced local smoothness) to ensure broad generalizability \cite{ge2024gated,li2023direction}.
Building on this analysis, as shown in Fig. \ref{fig:OverallofArch}, we design TenRPCANet, an architecture aligned with these intrinsic properties. However, the optimal way to construct such low-rank and sparse decomposition often varies across different tasks \cite{10496551,11015692,7508483}, and the non-uniqueness of tensor rank definitions \cite{9585029,10153488,8066348} makes explicit rank optimization problematic. 
To mitigate these limitations, rather than relying on deep unfolding methods \cite{10601492,joukovsky2023interpretable}, we directly exploit the intrinsic low-rank structure of video data by leveraging both local and non-local self-similarity through a self-attention mechanism \cite{11104998,9782722}, implicitly encoding low-rank priors without committing to a specific tensor formulation.

\begin{figure}[!t]
	\centering
	\includegraphics[width=1\linewidth]{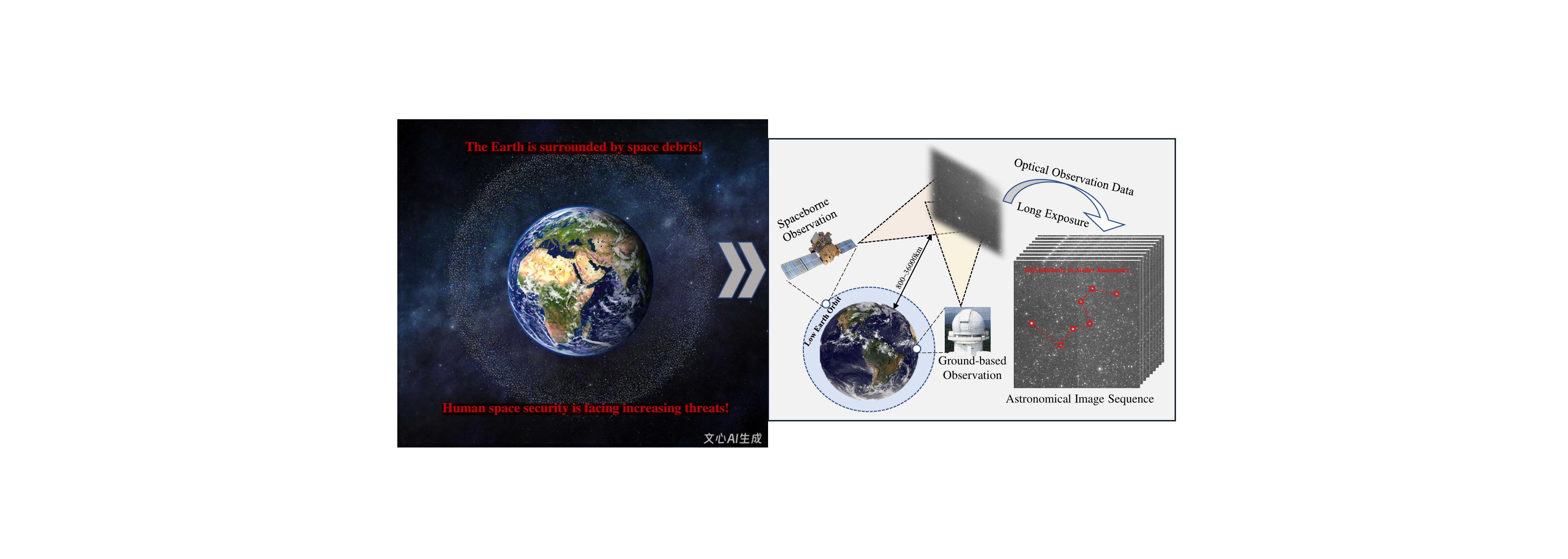}
	\caption{With the increasing frequency of space activities, the number of space debris in Earth's low Earth orbit (LEO) has grown dramatically. These debris typically travel at an average collision velocity of nearly 10 km/s. When impacting satellites, such high-speed collisions can cause irreversible damage or even lead to complete satellite failure, posing significant challenges to the safety and sustainability of space missions \cite{steindorfer2025space}.}
	\label{fig:RSO}
\end{figure}

To validate the effectiveness and generalizability of our method, we conduct experiments on two highly distinct and challenging tasks: multi-frame infrared small target detection and space debris detection (see Fig.~\ref{fig:RSO}). The former involves identifying dim targets under strong background clutter, while the latter requires distinguishing debris from millions of stars in astronomical images \cite{wang2024robust}. These tasks differ significantly in target appearance and motion \cite{chen2024convolutional}, making them ideal for assessing cross-task generalization.
Extensive evaluations on public datasets show that our method achieves robust performance under low signal-to-noise ratios and strong background interference. In contrast to methods that struggle to generalize across domains, our approach delivers consistent and transferable results in both scenarios.

The main contributions of this work can be summarized as follows:
\begin{itemize}
	\item We introduce a novel deep learning paradigm for small moving target detection, founded on the dual modeling principle that foreground localization and background suppression are inherently interdependent and mutually reinforcing.
	\item We conduct a physically grounded analysis of the background, foreground, and noise characteristics, relying on mild assumptions (PSF-induced local smoothness) to ensure broad applicability.
	\item Building upon these insights, we design TenRPCANet, a highly interpretable architecture that leverages spatiotemporal local and non-local self-similarity to represent the low-rank structure of video data.
	\item Extensive experiments on multi-frame infrared small target detection and space debris detection tasks validate the effectiveness and generalizability of our method.
\end{itemize}

The remainder of this paper is organized as follows. Section~\ref{Section:Related_Work} reviews related work. Section~\ref{Section:Method} presents our physically grounded analysis and introduces the method developed based on these insights. Experimental results and discussions are provided in Section~\ref{Section:Experiment}. Finally, Section~\ref{Section:Conclusion} concludes the paper.

\section{Related Work} \label{Section:Related_Work}
\subsection{Multi-Frame Infrared Small Target Detection}
Single-frame infrared small target detection \cite{gao2013infrared,9217948,DNANet,UIUNet,CSRNet,liu2023infrared,11080263,11146868,,11537388, MDCENet,HFMNet,WaveTD,DFAwareNet} often struggles in complex scenarios due to the semantic ambiguity of the targets and the lack of auxiliary information such as color and texture in the infrared modality. To address these challenges, multi-frame infrared small target detection has been proposed to exploit temporal information for enhanced detection performance.
Many existing methods adopt detector-based approaches \cite{11130659,10381806}; however, the center of the predicted bounding box often does not align with the actual center of the target, which poses challenges for accurately localizing small targets. Moreover, many downstream tasks, such as sequence unmixing \cite{11080063}, rely on the masks obtained from detection results.
Compared with model-driven approaches \cite{liuT2023infrared}, deep learning methods \cite{li2023direction,ying2025infrared,li2025probing} have achieved remarkable progress in segmentation-based multi-frame infrared small target detection. However, these methods often rely on motion cues and the structural sparsity of the targets, resulting in limited robustness and poor generalization to other tasks. 

In contrast to these studies, our core insight is that target detection and background discrimination are two sides of the same coin. The low-rank nature of the background remains consistent across various complex scenarios, making it more robust to identify targets by distinguishing the background. Furthermore, by relying on only mild, physically grounded assumptions (PSF-induced local smoothness), the proposed approach is more transferable to related tasks.
\subsection{Space Debris Detection}
In recent years, space debris detection utilizing spaceborne observation platforms has garnered increasing attention and become a prominent area of research \cite{11029687}. The dominant technical paradigm remains traditional model-driven approaches, which typically rely on the detect-before-track (DBT) framework \cite{lin2021new}. These methods first extract thousands of candidate targets from individual frames \cite{wang2025anomalous}, followed by multi-frame trajectory association to identify true targets \cite{9185021}. However, this pipeline is often complex \cite{cite-spmht,cite-stmht,cite-tmqht}, involving multiple stages with numerous hyperparameters \cite{schuckman2022using}. Moreover, it is generally difficult for such methods to simultaneously handle both streak-like and point-like targets under different operational modes \cite{zingarelli2014improving}, such as star tracking and target tracking. In recent years, some researchers have begun exploring deep learning-based approaches \cite{cite-sdebrisnet}. These methods heavily rely on structural sparsity priors and motion cues of the targets \cite{li2023direction}. However, the LSTM-based models they employ often struggle to effectively exploit temporal information \cite{chen2024convolutional}. Moreover, these methods typically require initialization during the detection process, which often leads to the exclusion of the initial frames.

Unlike the aforementioned methods, we leverage the structural consistency of stars, specifically their stable geometric relationships across frames, to reformulate space debris detection within a low-rank and sparse decomposition framework. We further employ a self-attention mechanism to capture spatiotemporal self-similarity among stars and explicitly model the consistency of topological relationships across frames to enhance star discrimination.
\subsection{Tokenization Strategies in Vision Transformers}
Early studies adopted a straightforward approach of extracting non-overlapping patches, flattening them, and applying embeddings \cite{dosovitskiy2020vit}; however, this procedure often led to instability during optimization. Extensive research has demonstrated that introducing early convolutional layers can substantially improve the inductive bias of Vision Transformers, enabling them to better capture local structures in visual data \cite{NEURIPS2021_ff1418e8}. This insight has led to the emergence of a variety of Conv-Stem architectures, such as HR-Stem \cite{10475592} and MSPE \cite{10962317}, which aim to enhance early-stage feature extraction. Beyond architectural innovations, several Conv-Stem variants have also been integrated into downstream tasks \cite{10124835,11091531}, consistently yielding notable performance gains.
However, most of these designs remain largely heuristic in nature \cite{khan2023survey}. In particular, they tend to emphasize either spatial-channel features or spatial-channel-temporal representations, often without a unified modeling framework. To compensate for limited theoretical grounding, these methods typically rely on the insertion of nonlinear activation functions within the stem to increase representational capacity \cite{Wang_Wang_Luo_Zhou_Zhou_Wang_Li_Jin_2022}.

Unlike prior methods, we propose Locally Subspace Embedding (LSE), which embeds third- and fourth-order tensor low-rank priors \cite{9115254,9780890,8618436,10742301,9730793} via a dual-branch design: a 3D patch branch approximates the fourth-order tensor construction by encoding local spatiotemporal structure, while a 2D multi-scale branch preserves local spatial structure. All operations in LSE are linear, omitting activation functions to avoid unnecessary nonlinear distortions. This design preserves the original relationships among pixels and patches, ensuring that the subsequent self-attention operates on a faithful representation of the video data.

\section{Methodology} \label{Section:Method}
We begin this section with a physically grounded analysis of the intrinsic characteristics of the background, foreground, and noise. Guided by the resulting priors, we develop TenRPCANet, a model tailored to leverage these structural properties.

\subsection{Analysis of the Intrinsic Properties of the Background} \label{Sec:Background}
To analyze the intrinsic priors of the background in the small moving target detection task and to motivate the appropriateness of both third-order \cite{9115254,9780890} and fourth-order \cite{8618436,10742301,9730793} low-rank tensor modeling strategies, which we argue are complementary in nature, we begin by examining two key empirical observations.
\begin{phenomenon}
\textit{It is well-known that, video data inherently possesses structural information and multidimensional redundancies.}
\end{phenomenon}
\noindent Video data intrinsically contains rich multidimensional structural information and redundancy \cite{6820779}, which can be formally characterized as spatial local correlations and non-local self-similarities \cite{8000407}. Specifically, let a video sequence be represented as a tensor 
\begin{equation}
	\boldsymbol{\mathcal{V}} \in \mathbb{R}^{H \times W \times T},
\end{equation}
where \(H, W, T\) denote the spatial height, spatial width, and temporal length respectively. To effectively capture the underlying low-rank structure, a natural approach is to extract overlapping spatiotemporal patches of size \(h \times w \times t\), i.e.,
\begin{equation}
	\boldsymbol{\mathcal{P}}_i = \boldsymbol{\mathcal{V}}[x_i : x_i+h-1,\, y_i : y_i+w-1,\, z_i : z_i+t-1] \in \mathbb{R}^{h \times w \times t}, \label{Eq:patch}
\end{equation}
where \((x_i, y_i, z_i)\) indicates the spatial-temporal coordinates of the patch, and \(i = 1, \ldots, P\) indexes the extracted patches. Due to the inherent non-local self-similarity in video data, patches \(\{\boldsymbol{\mathcal{P}}_i\}\) that exhibit similar structural patterns across different spatial locations can be grouped and stacked along an additional mode to form a fourth-order tensor \cite{8618436,10742301,9730793}:
\begin{equation}
	\boldsymbol{\mathcal{X}} \in \mathbb{R}^{h \times w \times t \times P}.
\end{equation}
Here, the first three modes \((h, w, t)\) represent the local spatiotemporal content of each patch, encoding strong local correlations, while the fourth mode aggregates non-local self-similar patches across the video volume, capturing long-range redundancies. This tensorial construction thus inherently leverages both local smoothness and non-local repetitive structures.

Importantly, the background in small moving target detection typically exhibits high spatiotemporal redundancy and regularity, which manifests as pronounced low-rankness in the tensor \(\boldsymbol{\mathcal{X}}\). Formally, the low-rank property can be expressed as
\begin{equation}
	\mathrm{rank}(\boldsymbol{\mathcal{X}}) \ll \min(h w t, P),
\end{equation}
providing a principled basis for discriminating background from sparse, anomalous moving targets. Consequently, modeling video data with such a fourth-order tensor effectively integrates the spatial, temporal, and patch similarity dimensions, forming a robust framework for background modeling in complex dynamic scenes.

\begin{figure}
	\centering
	\includegraphics[width=1.0\linewidth]{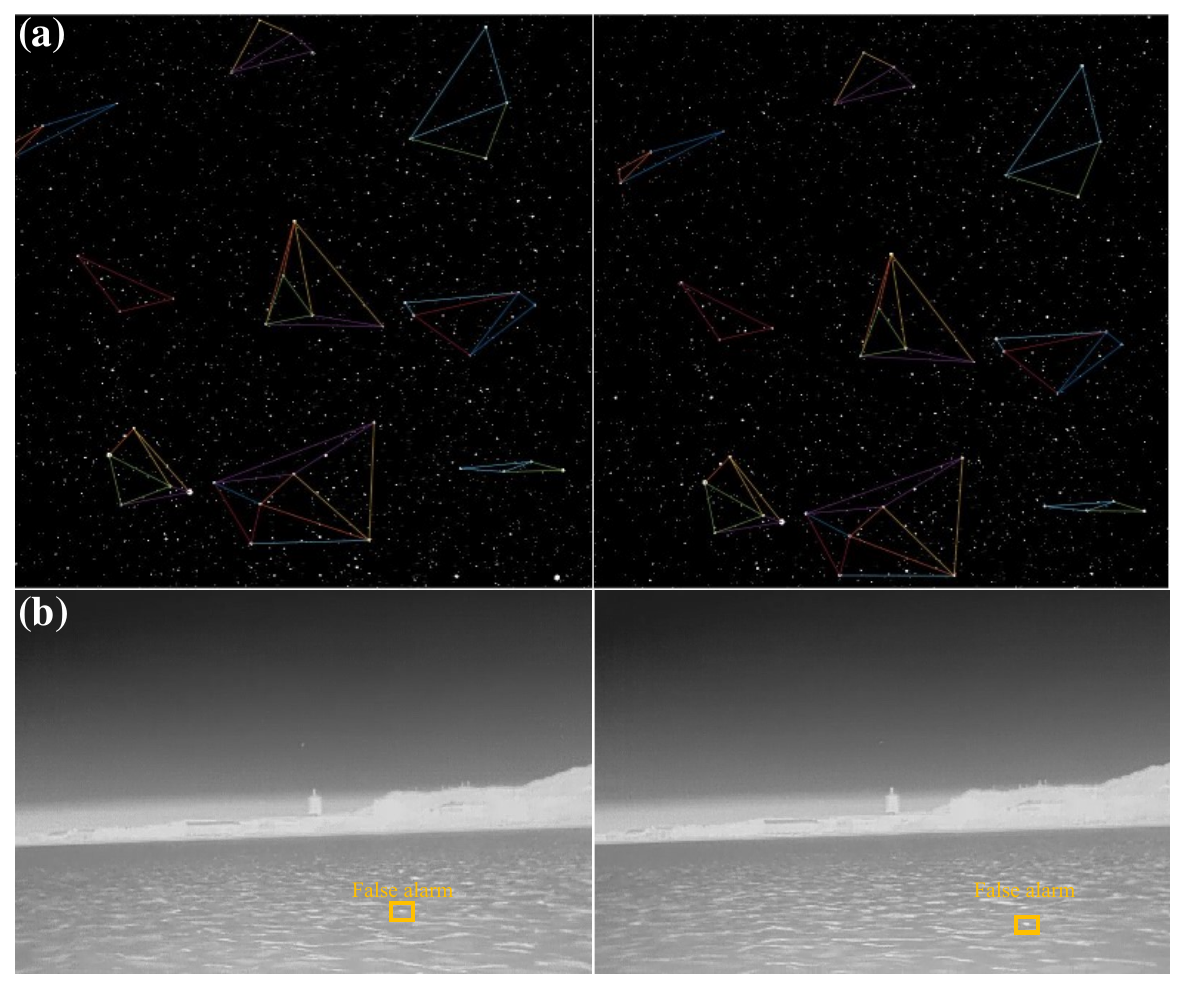}
	\caption{An important phenomenon is that false alarm sources may exhibit motion, yet they maintain relatively stable context. (a) In the space debris detection task, the stars have relatively stable topological relationships. We randomly sampled the topological relationships between some stars, and it is visually apparent that, despite significant displacements between frames, the triangular configurations remain stable. (b) In the multi-frame infrared small target detection task, false alarm sources also exhibit relatively stable context.}
	\label{fig:Star}
\end{figure}

\begin{phenomenon}
	\textit{False alarm sources typically exhibit relatively stable contextual patterns over time, whereas the context surrounding true targets tends to be unstable due to the targets’ inherent motion.}
\end{phenomenon}
\noindent As shown in Fig. \ref{fig:Star}, despite differences in imaging modalities, both infrared small target detection and space debris detection share a common prior. In the infrared domain, false alarms often arise from heterogeneous background structures or man-made objects with persistent appearance across frames. In space-based scenarios, false alarms are mainly caused by static background stars, whose apparent motion stems from platform drift but whose relative configuration remains unchanged~\cite{lin2023registration}. In contrast, true targets such as space debris do not share such stable contextual patterns, resulting in transient appearances.

While the fourth-order tensor construction \(\boldsymbol{\mathcal{X}} \in \mathbb{R}^{h \times w \times t \times P}\) captures both local spatiotemporal correlations and non-local self-similarities, it is insensitive to small-scale motion. Although this representation excels at reconstructing the background and suppressing noise, it cannot effectively detect candidate targets or determine them based on contextual information. Therefore, it is necessary to adopt a more conservative strategy that decouples spatial and temporal modeling. Specifically, we extract overlapping \(h \times w\) spatial patches from each frame independently and stack them into a third-order tensor \cite{9115254,9780890}:
\begin{equation}
	\boldsymbol{\mathcal{X}} = \mathrm{Stack} \left( \left\{ \boldsymbol{P}^{(t)}_i \right\} \right) \in \mathbb{R}^{h \times w \times N},
\end{equation}
where \(N\) is the total number of patches across all frames. This representation aggregates spatial appearance patterns across frames without strict temporal alignment, thereby enhancing robustness against minor background fluctuations and structured noise.

\subsection{Analysis of the Intrinsic Properties of Small Moving Targets} \label{Section:Target}

The concept of structured sparsity $J_{\boldsymbol{\mathcal{S}}}(\boldsymbol{\mathcal{S}})$ reflects that small targets, while sparse in their spatial distribution, inherently exhibit consistent visual structures, such as compact support and smooth intensity profiles. These characteristics are fundamentally influenced by the underlying physical image formation process. In particular, the imaging properties of small targets are significantly shaped by the optical system’s point spread function (PSF), especially in long-range observation scenarios. 
In ideal diffraction-limited systems, the PSF follows the Airy pattern derived from Fraunhofer diffraction theory~\cite{zhang2024dtnet}:
\begin{equation}
	\mathrm{PSF}(r) = \left[ \frac{2 J_1(\pi D r / \lambda f)}{\pi D r / \lambda f} \right]^2,
\end{equation}
where \(J_1(\cdot)\) is the first-order Bessel function, \(D\) is the aperture diameter, \(\lambda\) the wavelength, and \(f\) the focal length. This leads to a spatially compact, smooth intensity profile centered on the target \cite{zhang2024dtnet}.
Moreover, long-exposure imaging and relative motion introduce motion blur, yielding a combined degradation model \cite{lin2021new}:
\begin{equation}
	I(x, y) = \left[O(x, y) \otimes \mathrm{PSF}(x, y)\right] \otimes \mathcal{M}(x, y),
\end{equation}
where \(\mathcal{M}(x, y)\) denotes the motion blur kernel. Despite variability in target dynamics and sensor conditions, the resulting observations consistently exhibit two key properties: \textit{local intensity smoothness}, imposed by diffraction and motion blur, and \textit{spatial compactness}, due to the small physical size of the targets.
These physically grounded priors motivate detection strategies that emphasize spatial structure and appearance consistency, rather than relying primarily on potentially unreliable temporal dynamics~\cite{9098052}.

\subsection{Analysis of Noise Characteristics} \label{Section:Noise}
In practical applications, noise may exhibit structured patterns that neural networks can potentially exploit \cite{10507231,11029687}. In contrast, we focus on general, unstructured noise without semantic consistency or learnable patterns. Let \(\boldsymbol{\mathcal{V}}\), \(\boldsymbol{\mathcal{Y}}\), and \(\boldsymbol{\mathcal{N}}\) denote the observed video tensor, clean video, and additive noise respectively, modeled as
\begin{equation}
	\boldsymbol{\mathcal{V}} = \boldsymbol{\mathcal{Y}} + \boldsymbol{\mathcal{N}},
\end{equation}
where \(\boldsymbol{\mathcal{N}}\) follows an unknown, potentially complex distribution.
Although our unified model in Eq.~\eqref{Eq:forma} imposes an explicit regularization term $J_{\boldsymbol{\mathcal{N}}}(\boldsymbol{\mathcal{N}})$ to suppress noise, in practice the noise distribution is often unknown and complex. To address this, we adopt a Bayesian framework~\cite{10433564} and formulate the recovery of the clean video \(\boldsymbol{\mathcal{Y}}\) as a MAP estimation problem~\cite{5339210}:
\begin{equation}
	\boldsymbol{\mathcal{Y}}^\ast = \arg\max_{\boldsymbol{\mathcal{Y}}} \, p(\boldsymbol{\mathcal{Y}} \mid \boldsymbol{\mathcal{V}}) = \arg\max_{\boldsymbol{\mathcal{Y}}} \, p(\boldsymbol{\mathcal{V}} \mid \boldsymbol{\mathcal{Y}}) \cdot p(\boldsymbol{\mathcal{Y}}),
\end{equation}
where \(p(\boldsymbol{\mathcal{V}} \mid \boldsymbol{\mathcal{Y}})\) denotes the likelihood term, and \(p(\boldsymbol{\mathcal{Y}})\) encodes prior assumptions about the clean video, $p(\boldsymbol{\mathcal{Y}} \mid \boldsymbol{\mathcal{V}}) \propto p(\boldsymbol{\mathcal{V}} \mid \boldsymbol{\mathcal{Y}}) \cdot p(\boldsymbol{\mathcal{Y}}).$
To facilitate optimization, we take the negative logarithm of the posterior and obtain:
\begin{equation}
	\boldsymbol{\mathcal{Y}}^\ast = \arg\min_{\boldsymbol{\mathcal{Y}}} \, -\log p(\boldsymbol{\mathcal{V}} \mid \boldsymbol{\mathcal{Y}}) - \log p(\boldsymbol{\mathcal{Y}}).
\end{equation}
Given the unknown noise, the prior \(p(\boldsymbol{\mathcal{Y}})\) plays a key role \cite{5339210}. Since the clean video can be decomposed as $\boldsymbol{\mathcal{Y}} = \boldsymbol{\mathcal{L}} + \boldsymbol{\mathcal{S}}$, we do not need to impose overly restrictive assumptions on the distribution of $\boldsymbol{\mathcal{N}}$. Instead, we follow the consensus in model-driven infrared small target and space debris detection that noise is unstructured and does not persist across multiple frames.

\subsection{Overview of the Proposed TenRPCANet}
\begin{figure*}[!t]
	\centering
	\includegraphics[width=1\linewidth]{detailed_arch.pdf}
	\caption{Detail of the proposed TenRPCANet is designed based on our theoretical framework and consists of three core components. First, the Locally Subspace Embedding (LSE) introduces a novel tokenization strategy that implicitly regularizes subsequent self-attention mechanisms. Second, the Encoder and Decoder, constructed from Video Swin Transformer blocks, utilize spatiotemporal self-attention to effectively capture the low-rank structure of the background. Third, the Progressive Feature Refinement (PFR) Module, inspired by sparse tensor updates in Tensor RPCA, progressively refines the target features with minimal prior constraints (only PSF-Function).}
	\label{fig:OverallofDetailArch}
\end{figure*}
Motivated by the preceding analysis of background structure, target characteristics, and noise behavior, we propose TenRPCANet, a unified detection framework. Its overall architecture is illustrated in Fig.~\ref{fig:OverallofArch}. The input feature map has a shape of $B \times T \times C \times H \times W$, where $B$, $T$, $C$, $H$, and $W$ denote the batch size, temporal length, number of channels, height, and width, respectively. All frames within the temporal window are processed in parallel, and segmentation results for all $T$ frames are generated in a single forward pass.
Both encoder and decoder are designed to exploit low-rank structures, leveraging non-local self-similarity inherent in such representations. This enables effective modeling of low-rank priors without iterative optimization. To this end, Video Swin Transformer (VST) blocks \cite{liu2022video} are embedded in both encoder and decoder.

\subsection{Locally Subspace Embedding Module}
Our LSE module is designed to embed third- and fourth-order tensor low-rank priors into the self-attention mechanism while preserving the intrinsic structure of video data. This is achieved via a structure-preserving tokenization strategy.

To enforce low-rank priors on background structures modeled by third- and fourth-order tensors \cite{9115254,9780890,8618436,10742301,9730793}, we treat the tokenization strategy \(\mathcal{E}\) as an \textit{implicit regularizer} that maintains the geometric relationships among tokens. Specifically, \(\mathcal{E}\) defines an embedding manifold \(\mathcal{M} = \{\mathbf{z}_i = \mathcal{E}_i(\mathbf{X})\} \subset \mathbb{R}^d\), which governs the self-attention operator \cite{dosovitskiy2020vit}:
\begin{equation}
	\mathbf{SA}_i = \sum \alpha_{ij} \mathbf{v}_j\ \text{where}\ \alpha_{ij} = \frac{\exp(\mathbf{q}_i^\top \mathbf{k}_j)}{\sum_l \exp(\mathbf{q}_i^\top \mathbf{k}_l)},
\end{equation}
where \(\mathbf{q}_i = \mathbf{W}_q \mathbf{z}_i\), \(\mathbf{k}_j = \mathbf{W}_k \mathbf{z}_j\).
Assuming \(\mathcal{M}\) is a smooth manifold, the local inner products approximate geodesic distances \cite{meilua2024manifold}:
\begin{equation}
	\mathbf{q}_i^\top \mathbf{k}_j \approx -\frac{1}{2} d_{\mathcal{M}}^2(\mathbf{z}_i, \mathbf{z}_j), \label{Eq:geo}
\end{equation}
with \(d_{\mathcal{M}}\) the geodesic metric induced by \(\mathcal{E}\). Consequently, the attention weights correspond to a heat kernel on \(\mathcal{M}\) \cite{chen2023primal}:
\begin{equation}
	\alpha_{ij} \approx \exp\left(-\frac{d_{\mathcal{M}}^2(\mathbf{z}_i, \mathbf{z}_j)}{2\sigma^2}\right),
\end{equation}
and the self-attention operator acts as a diffusion process \cite{lafferty2005diffusion}:
\begin{equation}
	\mathbf{A}f(i) = \sum \alpha_{ij} f(j) \approx e^{-\sigma^2 \Delta_{\mathcal{M}}} f(i), \label{Eq:diff}
\end{equation}
where \(f(i)\) denotes the feature at point \(i\) on \(\mathcal{M}\), and \(\Delta_{\mathcal{M}}\) is the associated Laplace-Beltrami operator \cite{lafferty2005diffusion}. 
This formulation introduces a geometric inductive bias through \(\mathcal{E}\), implicitly regularizing self-attention without modifying its parametric form. 
By keeping the embedding faithful to the original pixel-space relationships, the self-attention operator respects the intrinsic geometry of the background manifold, preventing overfitting to target-specific patterns.

\noindent\textbf{Implementation Details.} The dual-branch design described below realizes this structure-preserving tokenization strategy by embedding complementary low-rank priors without distorting the original relationships among pixels. Specifically, the 3D patch branch approximates the fourth-order tensor patch construction described in Eq.~\eqref{Eq:patch}, thereby embedding local spatial-temporal structures into a compact representation, processes the input feature map $\mathbf{X} \in \mathbb{R}^{B \times C \times T \times H \times W}$ as follow:
\begin{equation}
	\mathbf{Y} = \phi_{3\times4\times4}(\mathbf{X}, \mathrm{Stride}=(1, 4, 4)), 
\end{equation}
where $\phi_{3 \times 4 \times 4}$ is a 3D convolution operator with kernel size $3 \times 4 \times 4$. This operation implicitly captures local spatial features without explicitly constructing overlapping patches, significantly reducing computational complexity.

In the 2D patch construction branch, we pursue two objectives: (i) introducing a third-order tensor low-rank prior to regularize the self-attention mechanism, and (ii) capturing contextual dependencies of false alarm sources. As shown in Eq.~\eqref{Eq:geo} and ~\eqref{Eq:diff}, an embedding that preserves local neighborhood relationships induces a more consistent geodesic metric, thereby allowing self-attention to accurately reflect the similarity structure of the input. To encourage such structure-preserving properties, we incorporate traditional multilinear manifold dimensionality reduction methods \cite{6021368,4032832}, which leverage low-rank tensor priors to preserve local linearity and mitigate geometric distortion. Building upon this, we design a Locally Linear Embedding (LLE) module that explicitly maintains neighborhood structures in a geometry-aware fashion. For an input feature map $\mathbf{X}$: 
\begin{equation}
	\mathbf{Z} = \psi_{1}(\mathrm{Concatenate}(\psi_{3}(\mathbf{X}),\psi_{5}(\mathbf{X}),\psi_{7}(\mathbf{X}))),
\end{equation}
where $\psi_k$ denotes a 2D convolution operator with kernel size $k \times k$, the output feature map is then downsampled. The resulting multi-scale feature map from the 2D patch construction branch is then downsampled and passed to the Progressive Feature Refinement (PFR) module.
Finally, the outputs from the two branches are fused via element-wise addition, integrating spatial-temporal cues for downstream encoding and decoding.

\begin{figure}
	\centering
	\includegraphics[width=1.0\linewidth]{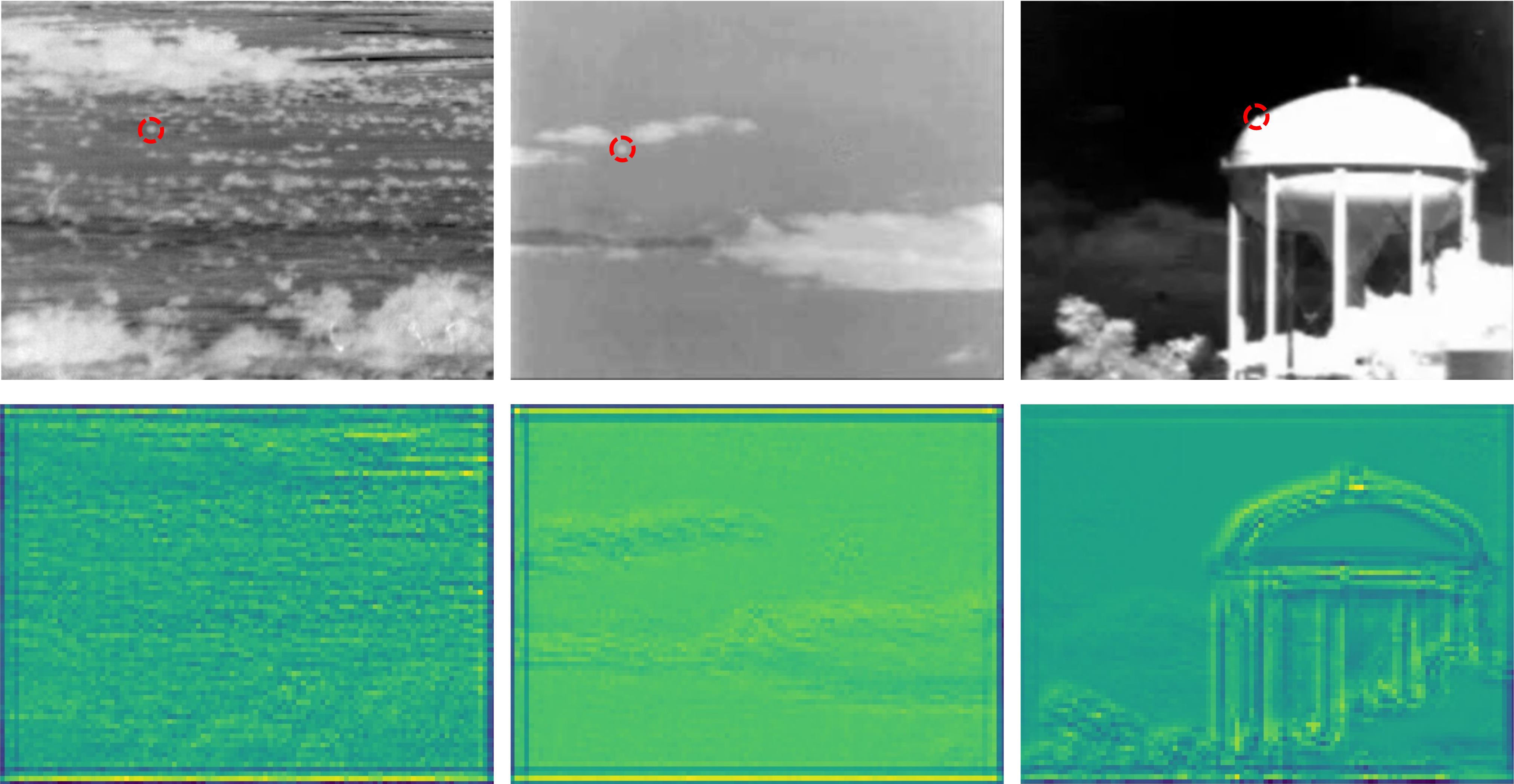}
	\caption{Visualization of the impact of the LSE module. The proposed design preserves the background structure instead of overemphasizing the target region (red circle), thereby maintaining the inherent relationships among pixels and patches and facilitating effective self-attention on faithful video representations.}
	\label{fig:LSE}
\end{figure}
The spatial downsampling (stride 4) in the LSE module significantly attenuates small target signals, which typically occupy only a few pixels. As a result, the feature maps passed to the VST contain primarily background structures (as shown in Fig. \ref{fig:LSE}); the encoder-decoder pathway thus models the low-rank background rather than tracking moving targets.

\noindent\textbf{Discussion.} We intentionally retain a linear projection in LSE instead of introducing additional nonlinear transformations. This design is motivated by several considerations. First, it is consistent with the theoretical analysis in Sec. \ref{Sec:Background}, where the low-rank property is derived from the intrinsic geometry of the input representation rather than from increased model expressiveness. Introducing unnecessary nonlinear mappings before tokenization would alter the local neighborhood structure assumed in our manifold analysis, making the theoretical interpretation less faithful. Second, classical model-driven approaches to video background modeling have consistently shown that temporal correlations among neighboring frames are predominantly linear or locally linear \cite{6122031}, as background variations are generally smooth while foreground targets occupy only a negligible portion of the scene. Preserving this linearity therefore provides a more appropriate inductive bias for background representation. Third, a linear embedding better preserves the geometric relationships inherited from the original pixel space \cite{6021368}, allowing neighboring tokens to maintain their relative similarity after tokenization, which is essential for subsequent self-attention to capture the underlying low-rank manifold. Finally, although video datasets often contain a large number of frames, their scene diversity is typically much lower than that of large-scale image datasets \cite{zhang2026matting}. Avoiding excessive nonlinear transformations before the Transformer therefore helps reduce the tendency to fit dataset-specific appearance variations while preserving the underlying background structure.

\subsection{Progressive Feature Refinement Module}
The decomposition $\boldsymbol{\mathcal{X}} = \boldsymbol{\mathcal{L}} + \boldsymbol{\mathcal{S}} + \boldsymbol{\mathcal{N}}$  is not uniquely identifiable without additional assumptions. In this work, we adopt two physically grounded assumptions to resolve this ambiguity: (1) targets exhibit local intensity smoothness due to the point spread function (Section \ref{Section:Target}); (2) noise is unstructured and does not persist across multiple frames, a common assumption in small target detection. 
In fact, this process naturally aligns with the sparse component update step in Tensor RPCA \cite{7488247}:
\begin{equation}
\begin{aligned}
		& \min_{\boldsymbol{\mathcal{L}},\boldsymbol{\mathcal{S}}}\ \mathrm{rank}(\boldsymbol{\mathcal{L}})+\lambda\|\boldsymbol{\mathcal{S}}\|_1 \\
		& \mathrm{s.t.}\quad\|\mathfrak{M}(\boldsymbol{\mathcal{X}}-\boldsymbol{\mathcal{L}})-\mathcal{S}\|^2_F\leq\eta 
		\label{Eq:diss}
\end{aligned}
\end{equation}
where $\mathfrak{M}(\cdot)$ denotes observation projection operator, and $\|\cdot\|_1$ denotes $\ell_1$-norm. Under the framework of ADMM algorithm, the augmented Lagrangian function of Eq. (\ref{Eq:diss}) can be given as follows:
\begin{equation}\begin{aligned}
		 &\Gamma(\boldsymbol{\mathcal{L}},\boldsymbol{\mathcal{S}},\boldsymbol{\Lambda},\mu) =\mathrm{rank}(\boldsymbol{\mathcal{L}})+\lambda\|\boldsymbol{\mathcal{S}}\|_1\\
		&+\frac{\mu}{2}\|\mathfrak{M}(\boldsymbol{\mathcal{X}}-\boldsymbol{\mathcal{L}})-\boldsymbol{\mathcal{S}}\|_F^2  +\langle\boldsymbol{\Lambda},\mathfrak{M}(\boldsymbol{\mathcal{X}}-\boldsymbol{\mathcal{L}})-\boldsymbol{\mathcal{S}}\rangle
\end{aligned}\end{equation}
where $\boldsymbol{\Lambda}$ represents the Lagrange multiplier. Since the background component $\boldsymbol{\mathcal{L}}_t$ has been effectively estimated by the encoder-decoder pathway, it can be treated as fixed in the subsequent optimization. Accordingly, the focus shifts to refining the sparse component $\boldsymbol{\mathcal{S}}$ given a fixed $\boldsymbol{\mathcal{L}}_t$:
\begin{equation}
\begin{aligned}
		\boldsymbol{\mathcal{S}}^* &=\arg\min_{\boldsymbol{\mathcal{S}}}\lambda\|\boldsymbol{\mathcal{S}}\|_1 \\
		& +\frac{\mu}{2}\|\mathfrak{M}(\boldsymbol{\mathcal{X}}-\boldsymbol{\mathcal{L}}_t)-\boldsymbol{\mathcal{S}}\|_F^2+\langle\boldsymbol{\Lambda},\mathfrak{M}(\boldsymbol{\mathcal{X}}-\boldsymbol{\mathcal{L}}_t)-\boldsymbol{\mathcal{S}}\rangle \\
		& =\arg\min_T\frac{1}{2}\left\|\boldsymbol{\mathcal{S}}-\left(\mathfrak{M}(\boldsymbol{\mathcal{X}}-\boldsymbol{\mathcal{L}}_t)\right.+\frac{1}{\mu}\boldsymbol{\Lambda}_{t-1}\right\|^2_F\\
		&+\frac{\lambda}{\mu}\|\boldsymbol{\mathcal{S}}\|_1
\end{aligned}
\end{equation}
Once the background component $\boldsymbol{\mathcal{L}}_t$ is estimated and fixed by the encoder-decoder, the minimization over 
$\boldsymbol{\mathcal{S}}$ becomes independent of the specific low-rank constraint on $\boldsymbol{\mathcal{L}}$. The optimal 
$\boldsymbol{\mathcal{S}}^*$
can then be obtained via the following closed-form expression:
\begin{equation}
	\boldsymbol{\mathcal{S}}^*=\mathrm{Soft}_{\frac{\lambda}{\mu}}\left(\mathfrak{M}(\boldsymbol{\mathcal{X}}-\boldsymbol{\mathcal{L}}_t)+\frac{1}{\mu}\boldsymbol{\Lambda}_{t-1}\right), \label{Eq:SVT}
\end{equation}
where $\mathrm{Soft}_{\frac{\lambda}{\mu}}(\cdot)$ is the soft threshold operator. Considering that, under the Tensor RPCA setting, the discriminative target estimation becomes a nearly linearized operation once the background region is determined, and that such piecewise linearization demonstrates strong generalizability across different types of rank constraints. This motivates a lightweight linear design that approximates soft-thresholding behavior while preserving structural consistency with traditional Tensor RPCA frameworks. 

Notably, after applying Tensor RPCA, the sparse components are typically reconstructed into full-resolution foreground maps via overlapping patch aggregation. This process implicitly encodes local spatial priors, enhancing target continuity and coherence. Although only $\ell_1$-norm regularization is used, the reconstruction induces structured sparsity by favoring spatially contiguous activations, thereby improving the integrity of small targets.
\begin{equation}
	\boldsymbol{\mathcal{T}} = \mathrm{Aggregate}\left(\boldsymbol{P}^{(\boldsymbol{\mathcal{S}})}_i\right). \label{Eq:AGG}
\end{equation}
Here, $\boldsymbol{P}^{(\boldsymbol{\mathcal{S}})}_i$ denotes the $i$-th patch extracted from the sparse tensor $\boldsymbol{\mathcal{S}}$. Typically, the operator $\mathrm{Aggregate} (\cdot)$ denotes median pooling aggregation \cite{gao2013infrared}.
This allows for the enforcement of structural sparsity on the targets without imposing strong prior assumptions (\eg~Saliency \cite{6781644}, Continuity \cite{6216381}).

\noindent\textbf{Implementation Details.} Given an input feature map 
$\mathbf{Z} \in\mathbb{R}^{B \times T \times W_1\times H_1 \times C_1}$, it is first reshaped into $\mathbf{X} \in\mathbb{R}^{BT \times C_1\times W_1\times H_1}$ to facilitate subsequent processing.
The proposed Progressive Feature Refinement module is constructed by stacking 2D sparse feature refinement blocks and upsampling layers.
The processing pipeline of the 2D sparse feature refinement block is as follows. Given two input feature maps $\mathbf{X} \in\mathbb{R}^{BT \times C_i\times W_i\times H_i}$ and $\mathbf{Y} \in\mathbb{R}^{BT \times C_i\times W_i\times H_i}$, with $\mathbf{X}$ coming from the previous layer and $\mathbf{Y}$ from the proposed Locally Subspace Embedding module via a skip connection. To remain consistent with the element-wise target localization process outlined in Eq. (\ref{Eq:SVT}), we employ a lightweight $1\times1$ convolutional layer to implement this functionality.
\begin{equation}
	\mathbf{Z} = \psi_{1}(\mathrm{Concatenate}(\mathbf{X}, \mathbf{Y})),
\end{equation} 
where $\mathbf{Z} \in\mathbb{R}^{BT \times C\times W\times H}$, and $\psi_k$ denotes a 2D convolution operator with kernel size $k \times k$. Next, considering the \textit{local intensity smoothness} and \textit{spatial compactness} of small targets, we apply a simple $3\times3$ convolution to capture their local characteristics. Importantly, only linear operations are used in this step to remain consistent with the patch-based aggregation process in sparse component reconstruction (Eq.~\eqref{Eq:AGG}). It is worth noting that extensive empirical evidence from model-driven approaches suggests that incorporating locality-aware operators is effective in handling weak targets \cite{PR16MPCM}.
\begin{equation}
	\hat{\mathbf{Z}} = \psi_{3}(\mathbf{Z}) + \mathbf{Z}
\end{equation}
An upsampling module \cite{chen2024convolutional} is applied immediately after each 2D sparse feature refinement block. To approximate the nonlinear suppression behavior of soft-thresholding operator and promote sparsity in the output, we apply a Sigmoid activation followed by a confidence-based hard thresholding. The final output is computed as:
\begin{equation}
	\mathbf{Out} = \sigma(\hat{\mathbf{Z}}) \cdot \mathbb{I}\left[\sigma(\hat{\mathbf{Z}}) \geq \tau \right],
\end{equation}
where \( \sigma(\cdot) \) denotes the Sigmoid function, \( \tau \in (0,1) \) is a confidence threshold, and \( \mathbb{I}[\cdot] \) is the indicator function that outputs 1 when the condition is true and 0 otherwise. This formulation softly maps activations to the confidence domain while enforcing sparsity through binary gating, effectively mimicking the nonlinear suppression behavior of soft-thresholding. The gating mechanism mimics soft-thresholding behavior and effectively suppresses weak activations, thereby reducing false positives in practical detection scenarios.

\subsection{Loss Function}
The small moving target detection task can be formulated as a binary classification problem. We employ the binary cross-entropy loss as the training objective. The overall loss is computed as
\begin{equation}
	\mathcal{L}=\sum_{k=1}^T\sum_{j=1}^H\sum_{i=1}^W\mathcal{L}_\mathrm{BCE}\left(\mathbf{Pred}\left(i,j,k\right),\mathbf{GT}\left(i,j,k\right)\right),
\end{equation}
where $\mathbf{Pred}\left(i,j,k\right) \in (0, 1)$ and $\mathbf{GT}\left(i,j,k\right) \in \{0, 1\}$ denote the predicted confidence map and the ground truth at pixel $(i,j)$ in the $k$-th frame, respectively. Since this is a sequence-to-sequence process, both feature maps have the shape $T \times H \times W$.

\subsection{Discussion with Deep Unfolding Methods}
We highlight key differences from deep unfolding of tensor RPCA. First, the non-uniqueness of tensor rank forces deep unfolding to commit to a specific formulation, introducing a strong prior that may not generalize across tasks. Second, existing methods, such as ROMAN \cite{joukovsky2023interpretable}, rely on SVD during inference, which is computationally prohibitive for high-resolution inputs (\eg, $1024\times1024$). Third, to maintain sensitivity to small targets, deep unfolding methods typically construct overlapping patches (\eg, $50\times50$ with stride 15) \cite{gao2013infrared}, yielding thousands of patches (\eg, 4225 per frame), which is impractical even on an A100 GPU. Fourth, our framework jointly leverages third- and fourth-order tensor constructions, which capture complementary background properties but have fundamentally different structures; they cannot be unified into a single optimization objective suitable for deep unfolding. In contrast, our method avoids these limitations by implicitly encoding low-rank priors via self-attention, achieving better generalization with a simpler architecture, as validated by our cross-domain experiments.

\subsection{Theoretical Analysis of Low-Rank Self-Attention}
\label{sec:theoretical_analysis}
Let $\mathbf{X}\in\mathbb{R}^{M\times C}$ denote the input feature representation of the proposed LSE module, where $M=H\times W$ represents the number of spatial locations and $C$ denotes the feature dimension. Due to the strong spatial correlation and redundancy of infrared background regions, the feature representation can be approximately characterized by a
low-dimensional subspace, which is consistent with the low-rank background prior widely adopted in infrared small target detection. Therefore, we assume that
\begin{equation}
	{\rm rank}(\mathbf{X})=r_x,
	\quad
	r_x\ll {\rm min}(M,C).
\end{equation}
The proposed LSE module aggregates spatial features into a compact token representation. Specifically, the token aggregation operation in LSE can be formulated as
\begin{equation}
	\mathbf{Z}=\mathbf{P}\mathbf{X},
\end{equation}
where $\mathbf{P}\in\mathbb{R}^{N\times M}$ denotes the learned aggregation matrix and $\mathbf{Z}\in\mathbb{R}^{N\times C}$ represents the generated token representation.
\begin{theorem}[Low-rank boundedness of LSE-based self-attention]
	\label{thm:low_rank_attention}
	Given the input feature representation $\mathbf{X}$ and the token representation $\mathbf{Z}$ generated by the LSE module, the following relationship holds:
	\begin{equation}
		{\rm rank}(\mathbf{Q}\mathbf{K}^{T})
		\leq
		{\rm rank}(\mathbf{Z})
		\leq
		{\rm rank}(\mathbf{X}).
	\end{equation}
	where the query and key embeddings are obtained by
	\begin{equation}
		\mathbf{Q}=\mathbf{Z}\mathbf{W}_q,\quad
		\mathbf{K}=\mathbf{Z}\mathbf{W}_k,
	\end{equation}
	with $\mathbf{W}_q,\mathbf{W}_k\in\mathbb{R}^{C\times d}$ being learnable projection matrices. Here, $\mathbf{Q}\mathbf{K}^{T}$ denotes the	attention affinity matrix before the softmax normalization.
\end{theorem}
\begin{proof}
	We first analyze the rank relationship between the input representation	and the generated tokens. Since the LSE token aggregation operation can be written as
	\begin{equation}
		\mathbf{Z}=\mathbf{P}\mathbf{X},
	\end{equation}
	according to the rank inequality of matrix multiplication,
	\begin{equation}
		{\rm rank}(\mathbf{A}\mathbf{B})
		\leq
		\min({\rm rank}(\mathbf{A}),
		{\rm rank}(\mathbf{B})),
	\end{equation}
	we have
	\begin{equation}
		\begin{aligned}
			{\rm rank}(\mathbf{Z})
			&=
			{\rm rank}(\mathbf{P}\mathbf{X})\\
			&\leq
			\min({\rm rank}(\mathbf{P}),
			{\rm rank}(\mathbf{X}))\\
			&\leq
			{\rm rank}(\mathbf{X}).
		\end{aligned}
	\end{equation}
	Therefore, the token aggregation operation in LSE cannot increase the intrinsic rank of the input feature representation.
	Next, we analyze the rank property of the attention affinity matrix. The attention affinity matrix before softmax normalization is defined as
	\begin{equation}
		\mathbf{A}_s=
		\mathbf{Q}\mathbf{K}^{T}.
	\end{equation}
	According to the rank inequality,
	\begin{equation}
		{\rm rank}(\mathbf{A}_s)
		\leq
		\min({\rm rank}(\mathbf{Q}),
		{\rm rank}(\mathbf{K})).
	\end{equation}
	Since
	\begin{equation}
		\mathbf{Q}=\mathbf{Z}\mathbf{W}_q,\quad
		\mathbf{K}=\mathbf{Z}\mathbf{W}_k,
	\end{equation}
	we further have
	\begin{equation}
		\begin{aligned}
			{\rm rank}(\mathbf{Q})
			&=
			{\rm rank}(\mathbf{Z}\mathbf{W}_q)
			\leq
			{\rm rank}(\mathbf{Z}),\\
			{\rm rank}(\mathbf{K})
			&=
			{\rm rank}(\mathbf{Z}\mathbf{W}_k)
			\leq
			{\rm rank}(\mathbf{Z}).
		\end{aligned}
	\end{equation}
	Consequently,
	\begin{equation}
		{\rm rank}(\mathbf{Q}\mathbf{K}^{T})
		\leq
		{\rm rank}(\mathbf{Z}).
	\end{equation}
	Combining the above two inequalities gives
	\begin{equation}
		{\rm rank}(\mathbf{Q}\mathbf{K}^{T})
		\leq
		{\rm rank}(\mathbf{Z})
		\leq
		{\rm rank}(\mathbf{X}).
	\end{equation}
	This completes the proof.
\end{proof}
\noindent\textbf{Remark.} The theorem establishes that the attention affinity matrix \(\mathbf{QK}^T\) is low-rank. This low-rankness is explicitly constrained by the LSE module, since \(\operatorname{rank}(\mathbf{QK}^T) \leq \operatorname{rank}(\mathbf{Z}) \leq \operatorname{rank}(\mathbf{X})\). The encoder-decoder architecture is naturally motivated by the scale effect of non-local self-similarity in video backgrounds, where repetitive patterns exist across both local and distant regions and require hierarchical modeling. More importantly, the stride-4 downsampling in LSE substantially attenuates the high-frequency target signals, as visualized in Fig. \ref{fig:LSE}. Since small targets occupy only 1--3 pixels, they are largely eliminated before reaching the encoder. The self-attention mechanism is therefore compelled to model the remaining low-frequency background structures. This combination of algebraic rank bounds and physical signal attenuation guarantees that the encoder-decoder extracts a genuine low-rank background \(\mathcal{L}\) rather than overfitting to target textures or noise.

\section{Experiment}\label{Section:Experiment}
\subsection{Experimental Setup}
\subsubsection{Datasets} 
\begin{figure}
	\centering
	\includegraphics[width=\linewidth]{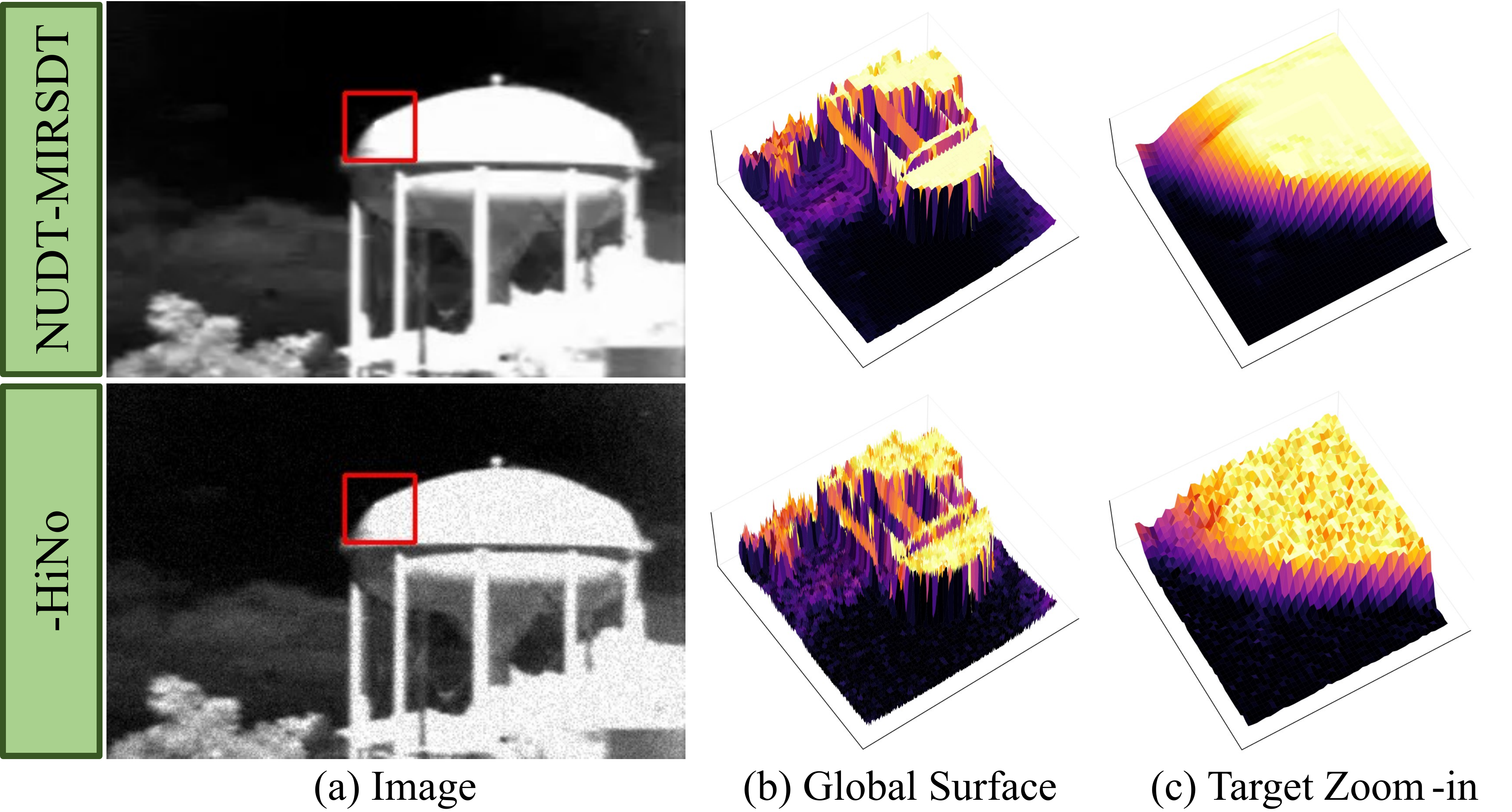}
	\caption{Representative examples from the NUDT-MIRSDT and NUDT-MIRSDT-HiNo datasets. In contrast to conventional infrared small target detection datasets, where most targets exhibit clear visual saliency, these datasets contain extremely weak targets with very low signal-to-background contrast, making them ideal for evaluating the perception capability of detection models under challenging conditions. Notably, even after local magnification, the targets remain barely discernible to the human eye.}
	\label{fig:IR}
\end{figure}

In this paper, we validate the effectiveness and generalization ability of the proposed TenRPCANet on two highly challenging downstream tasks: multi-frame infrared small target detection and space debris detection, which differ significantly in both target characteristics and motion patterns. For the multi-frame infrared small target detection task, we evaluate our method on two highly challenging benchmarks, NUDT-MIRSDT \cite{li2023direction} and NUDT-MIRSDT-HiNo \cite{li2025probing}, both of which feature complex cluttered backgrounds and low signal-to-noise ratios (as shown in Fig. \ref{fig:IR}). The NUDT-MIRSDT dataset \cite{li2023direction} contains 120 sequences, each consisting of 100 frames. The test set is divided into two subsets according to SNR: one with 8 sequences (SNR < 3) and the other with 12 sequences (SNR between 3 and 10), with an average SNR of 7. The NUDT-MIRSDT-HiNo dataset \cite{li2025probing} introduces strong structured noise based on the same sequences, following the same data split, and has an average SNR of 3.24 in the test set.

For the space debris detection task, we conduct experiments on the SSTD-S (Synthesis Set) and the SSTD-R (Real Optical Set) \cite{chen2024convolutional}. 
The SSTD-S dataset is generated by simulating physical imaging processes, producing targets with varying intensities, distributions, streak orientations, and lengths. Gaussian white noise is added to model background clutter, while target appearance and intensity profiles remain consistent across frames. Following the official protocol, the training set contains 750 sequences, and the test set contains 600 sequences. Each sequence consists of 20 frames of size \(1024\times1024\), with corresponding mask annotations.
The SSTD-R dataset is collected by a spaceborne observation platform in low Earth orbit. It includes 70 training sequences and 21 test sequences, each containing 10 to 30 frames of size \(1024\times1024\). Ground-truth annotations are provided by the Changchun Satellite Observatory based on a space debris observation database. The images have been pre-processed following standard space object processing pipelines to suppress stray light and smear effects.

\subsubsection{Evaluation Metrics} We follow standard evaluation protocols for each task. For multi-frame infrared small target detection \cite{ying2025infrared}, we adopt pixel-level metrics including the Probability of Detection ($P_d$), False Alarm rate ($F_a$), and the Area Under the Curve (AUC). For the space debris detection task \cite{chen2024convolutional}, we evaluate both object-level and pixel-level performance. Specifically, we report the Recall ($\mathrm{R}^t$), False Alarms ($\mathrm{FA}^t$), and $\mathrm{F1}^t$-score at the object level, along with the Intersection over Union (IoU) at the pixel level. Note that the final values are reported as sequence level averages.

\subsubsection{Implementation Details} The proposed method is trained on an NVIDIA A100 GPU using PyTorch 1.8.2 and CUDA 11.2. Adam optimizer with a learning rate schedule (ReduceLROnPlateau, 1e-4 to 1e-8) is used. Input normalization accounts for the bit depth and histogram characteristics of the data. Infrared images are normalized by 255.0 (8-bit), while optical astronomical images are normalized by 65535.0 (16-bit) to mitigate histogram truncation effects. The input resolutions are \(320\times416\) for infrared sequences and \(1024\times1024\) for space debris sequences.

\subsection{Comparison with State-of-the-Arts}

\begin{table*}[t!]
	\caption{Comparison of experiment results on the NUDT-MIRSDT dataset and the NUDT-MIRSDT-HiNo dataset. The best results are in \textbf{bold}, and the second-best results are \underline{underlined}. \textit{SF} and \textit{MF} refer to single-frame and multi-frame methods, respectively.}\label{tab:SOTA1}
	\centering
	\resizebox{\textwidth}{!}{
		\renewcommand{\arraystretch}{1.1}
		\begin{tabular}{c c l c c c c c c c c r c}
			\noalign{\hrule height 1pt}
			\multicolumn{3}{c}{\multirow{2}*{Methods}} & \multicolumn{2}{c}{$SNR\leq 3$} & \multicolumn{3}{c}{NUDT-MIRSDT} & \multicolumn{3}{c}{NUDT-MIRSDT-HiNo} & \multirow{2}*{Param (M)} &\multirow{2}*{FPS}\\
			\cline{4-11}
			\multicolumn{3}{c}{~}                    & $P_d$  & $F_a$        & $P_d$  & $F_a$ & AUC       & $P_d$  & $F_a$  & AUC & & \\ \noalign{\hrule height 1pt}
			\multirow{8}*{\rotatebox{90}{Traditional Methods}} & \multicolumn{1}{c}{\multirow{8}{*}{\rotatebox{90}{\textit{MF}}}}  & MSLSTIPT \cite{sun2020infrared} \textit{(TGRS'20)} & 4.16 & 21.70 & 18.97 & 15.37 & 0.9404 & 3.93 & 73.76 & 0.9185 & - & 0.17 \\
			& \multicolumn{1}{c}{}  & IMNN-LWEC \cite{luo2022imnn} \textit{(TGRS'22)} & 0.00 & 7.22 & 26.43 & 10.74 & 0.6734 & 4.97 & 83.28 & 0.5394 & - & 0.31 \\
			& \multicolumn{1}{c}{}  & SRSTT \cite{li2023sparse} \textit{(TGRS'23)} & 69.94 & 6.12 & 90.63 & 3.35 & \textbf{0.9989} & 4.34 & 55.04 & 0.5358 & - & 0.06 \\  
			& \multicolumn{1}{c}{}  & 4DST-BTMD \cite{luo20234dst} \textit{(TGRS'23)} & 41.58 & 23.45 & 44.77 & 74.95 & 0.8488 & 4.80 & 77.29 & 0.6651 & - & 26.09 \\  
			& \multicolumn{1}{c}{}  & STRL-LBCM \cite{luo2023spatial} \textit{(TAES'23)} & 5.48 & 85.53 & 19.03 & 34.05 & 0.5972 & 2.55 & 77.78 & 0.5238 & - & 0.87 \\  
			& \multicolumn{1}{c}{}  & 4D-TR \cite{wu2023infrared} \textit{(TGRS'23)} & 55.77 & 2.55 & 55.70 & 3.19 & 0.9946 & 4.63 & 120.16 & 0.6633 & - & 0.36 \\ 
			& \multicolumn{1}{c}{}  & 4D-TT \cite{wu2023infrared} \textit{(TGRS'23)} & 24.95 & 1.67 & 30.89 & 3.21 & 0.8287 & 6.94 & 73.18 & 0.5347 & - & 0.82 \\  
			& \multicolumn{1}{c}{}  & NFTDGSTV \cite{liuT2023infrared} \textit{(TGRS'23)} & 1.51 & 32.31 & 13.77 & 35.32 & 0.8613 & 11.56 & 43.16 & \underline{0.9524} & - & 0.58 \\ \hline  
			\multirow{18}*{\rotatebox{90}{Deep-Learning Methods}}  & \multicolumn{1}{c}{\multirow{12}{*}{\rotatebox{90}{\textit{SF}}}}  & ACM \cite{Dai_2021_WACV} \textit{(WACV'21)} & 7.75 & 22.88 & 51.533 & 17.52 & 0.9298 & 0 & - & 0.8727 & 0.398 & 57.89 \\
			& \multicolumn{1}{c}{}  & ALCNet \cite{ALCNet} \textit{(TGRS'21)} & 3.97 & 37.10 & 52.57 & 25.50 & 0.8435 & 36.09 & 91.99 & 0.9326 & 0.864 & 55.14 \\
			& \multicolumn{1}{c}{}  & Res-UNet \cite{xiao2018weighted} \textit{(ITME'18)} & 15.83 & 30.32 & 63.27 & 40.83 & 0.9198 & 35.51 & 22.55 & 0.9391 & \underline{0.227} & 77.06 \\
			& \multicolumn{1}{c}{}  & ISNet \cite{ISNet} \textit{(CVPR'22)} & 17.96 & 8.53 & 65.99 & 19.25 & 0.9123 & 28.40 & 90.17 & 0.9224 & 1.09 & 25.14 \\
			& \multicolumn{1}{c}{}  & UIUNet \cite{UIUNet} \textit{(TIP'23)} & 15.12 & 17.46 & 61.25 & 14.42 & 0.9436 & 43.67 & 28.87 & 0.9246 & 50.52 & 22.75 \\
			& \multicolumn{1}{c}{}  & DNA-Net \cite{DNANet} \textit{(TIP'23)} & 23.74 & 19.23 & 67.38 & 15.07 & 0.8843 & 49.16 & 60.98 & 0.9373 & 4.698 & 12.17 \\
			& \multicolumn{1}{c}{}  & AGPCNet \cite{AGPCNet} \textit{(TAES'23)} & 31.76 & 176.38 & 55.47 & 85.56 & 0.9443 & 42.452 & 13655.20 & 0.7348 & 12.36 & 13.05 \\
			& \multicolumn{1}{c}{}  & MSHNet \cite{MSHNet} \textit{(CVPR'24)} & 2.46 & 78.07 & 36.78 & 41.91 & 0.7966 & 21.81 & 44.92 & 0.6748 & 4.06 & 17.83 \\
			& \multicolumn{1}{c}{}  & SCTransNet \cite{SCTransNet} \textit{(TGRS'24)} & 23.63 & 122.24 & 62.81 & 74.20 & 0.9320 & 20.65 & 35.37 & 0.8283 & 11.32 & 10.07 \\
			& \multicolumn{1}{c}{}  & MiM-ISTD \cite{10740056} \textit{(TGRS'24)} & 0.00 & 71.24 & 15.27 & 50.12 & 0.9152 & 1.91 & 664.24 & 0.6879 & 8.59 & 25.41 \\
			& \multicolumn{1}{c}{}  & RPCANet \cite{RPCANet} \textit{(WACV'24)} & 30.06 & 81.21 & 61.13 & 41.28 & 0.8694 & 21.81 & 198.14 & 0.8786 & 0.68 & 7.90 \\
			& \multicolumn{1}{c}{}  & ILNet \cite{10899837} \textit{(TAES'25)} & 17.39 & 55.89 & 57.90 & 34.09 & 0.7572 & 34.11 & 56.53 & 0.6551 & 6.322 & 36.82 \\ \cline{2-13}  
			& \multicolumn{1}{c}{\multirow{6}{*}{\rotatebox{90}{\textit{MF}}}}
			& STDMANet \cite{yan2023stdmanet} \textit{(TGRS'23)} & 92.82 & 2.88 & 96.59 & 3.40 & 0.9908 & 51.65 & \underline{1.95} & 0.8766 & 11.88 & 5.16 \\ 
			& \multicolumn{1}{c}{}  & Res-U+RFR \cite{ying2025infrared} \textit{(TGRS'25)} & 64.65 & 24.09 & 88.61 & 11.58 & 0.9502 & 35.11 & 464.92 & 0.8655 & 1.02 & 34.77 \\ 
			& \multicolumn{1}{c}{} & Res-U+DTUM \cite{li2023direction} \textit{(TNNLS'25)} & 91.68 & 2.37 & 97.46 & 3.00 & 0.9967 & 43.90 & 4.86 & 0.9413 & 0.30 & 25.39 \\
			& \multicolumn{1}{c}{} & DQAligner \cite{11363482} \textit{(TGRS'26)} & 81.29 & \textbf{0.37} & 94.22 & 0.15 & 0.9419 & 33.41 & 2.95 & 0.8022 & 2.410 & 10.01 \\
			& \multicolumn{1}{c}{}  & DeepPro \cite{li2025probing} \textit{(TPAMI'26)} & \underline{95.84} & \underline{0.52} & \underline{98.50} & \underline{0.72} & 0.9973 & \underline{59.17} & \textbf{1.76} & \underline{0.9638} & \textbf{0.197} & \underline{155.40} \\ \rowcolor{gray!20}
			& \multicolumn{1}{c}{}  & TenRPCANet (Ours) & \textbf{98.53} & 1.3 & \textbf{99.33} & \textbf{0.36} & \underline{0.9978} & \textbf{86.62} & 6.37 & \textbf{0.9823} & 1.78 & \textbf{176.24} \\ \noalign{\hrule height 1pt}
		\end{tabular}
		}
\end{table*}
\begin{figure}
	\centering
	\includegraphics[width=0.85\linewidth]{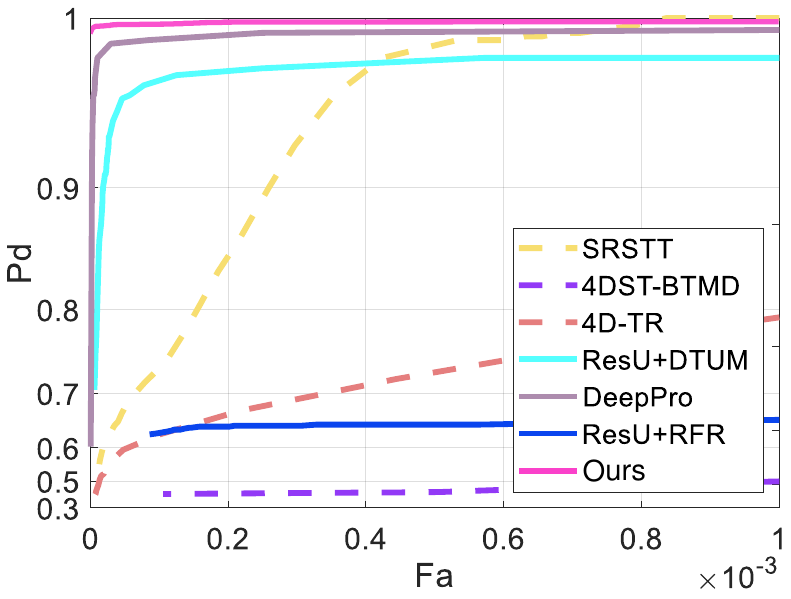}
	\caption{ROC curve on the NUDT-MIRSDT dataset ($SNR \leq 3$).}
	\label{fig:ROC}
\end{figure}
\begin{table*}[tbp]
	\centering
	\caption{Comparison of experiment results on synthetic data and real optical data. The best and second-best results for each metric are indicated in \textbf{bold} and \underline{underline}, respectively.}
		\resizebox{\textwidth}{!}{\begin{tabular}{l|cccc|cccc|ccc}
			\toprule
			\multirow{2}{*}{Method} & \multicolumn{4}{c|}{SSTD-S (Synthesis Set)\dag}   & \multicolumn{4}{c|}{SSTD-R (Real Optical Set)\ddag} & \multicolumn{3}{c}{Scale}\\
			\multicolumn{1}{c|}{}                        & $\text{R}^t$($\uparrow$)     & $\text{FA}^t$($\downarrow$)  & $\text{F1}^t$($\uparrow$) & IoU($\uparrow$)               
			& $\text{R}^t$($\uparrow$)     & $\text{FA}^t$($\downarrow$)  & $\text{F1}^t$($\uparrow$) & IoU($\uparrow$)   
			& Param (M)         & GFLOPs            & Time (ms)*
			\\
			\midrule
			\multicolumn{12}{l}{\textit{Traditional Methods}} \\
			\hline
			$\mathrm{TMQHT}_{16}$ \cite{cite-tmqht}
			& 84.52\%               & 0.355\%               & 87.02\%              & -                 
			& 83.69\%               & 6.125\%               & 86.32\%              & -
			& - & - & \phantom{0}1,435 \\ 
			$\mathrm{SPMHT}_{19}$ \cite{cite-spmht}
			& 33.01\%               & 4.134\%               & 36.64\%              & -                  
			& 70.52\%               & 26.16\%               & 72.14\%              & -
			& - & - & \phantom{00,}974 \\ 
			$\mathrm{STMHT}_{20}$ \cite{cite-stmht}
			& 52.83\%               & 0.243\%               & 54.59\%              & -                  
			& 71.02\%               & \underline{3.827\%}   & 76.07\%              & -
			& - & - & \phantom{0}1,756 \\ 
			\hline
			\multicolumn{12}{l}{\textit{Deep-Learning Methods}} \\
			\hline
			$\mathrm{DNANet}_{23}$ \cite{DNANet}
			& 76.63\%               & 0.256\%               & 78.87\%              & 41.45\%
			& 95.81\%               & 29.51\%               & 81.22\%              & 55.30\%
			& \phantom{00}4.70      & \phantom{0}228.6      & 11,712  \\
			$\mathrm{UIUNet}_{23}$ \cite{UIUNet}
			& 82.63\%               & \underline{0.219\%}   & 84.69\%              & 45.24\%            
			& 94.20\%               & 22.79\%               & 84.86\%              & 55.84\%
			& \phantom{0}50.54      & \phantom{0}872.6      & \phantom{0}1,564  \\
			$\mathrm{SDebrisNet}_{23}$ \cite{cite-sdebrisnet}
			& 75.76\%               & 1.013\%               & 78.48\%              & 56.52\%                 
			& 95.39\%               & 12.28\%               & 91.39\%              & 61.94\%
			& \phantom{00}1.69	    & \phantom{00}\bf{11.0} & \phantom{00,}\bf{386}   \\
			$\mathrm{DnTNet}_{24}$ \cite{chen2024convolutional}
			& 87.79\%               & 29.54\%               & 69.28\%              & \underline{59.33\%}
			& \underline{98.10\%}               & 8.903\%               & 94.47\%              & \underline{79.04\%}
			& \phantom{00}3.37      & \phantom{0}219.8	    & \phantom{00,}802   \\
			$\mathrm{LMAFormer}_{24}$ \cite{cite-lmaformer}
			& \underline{92.85\%}          & 3.739\%               & \bf{93.53\%}         & 53.60\%		
			& 97.67\%               & 4.883\%               & \underline{96.10\%}              & 38.06\%
			& 590.05                & 1946.3                & 22,754 \\
			$\mathrm{DTUM}_{25}$ \cite{li2023direction}
			&89.37\%               & 84.76\%               & 25.85\%              & 13.08\%
			& 84.86\%               & 61.42\%               & 51.11\%              & 39.59\%
			& \phantom{00}\underline{0.30} & \phantom{0}298.3      & \phantom{0}3,503 \\
			$\mathrm{DeepPro}_{26}$ \cite{li2025probing}
			& \bf{93.94\%}               & 3.678\%               & \underline{93.31\%}              & 60.17\%
			& 94.90\%               & 9.83\%               & 91.89\%              & 54.56\%
			& \phantom{0}\bf{0.197}     & \phantom{0}121.78      & \phantom{0}1756  \\
			\hline
			\multicolumn{12}{l}{\textit{Proposed Method}} \\ \hline \rowcolor{gray!20}
			TenRPCANet (Ours)
			& 87.81\%   & \bf{0.054\%}          & 89.73\%  & \bf{74.79\%}
			& \bf{99.53\%}          & \bf{1.000\%}          & \bf{99.00\%}         & \bf{80.78\%}
			& \phantom{00}1.78  & \phantom{00}\underline{79.26}  & \phantom{00,}\underline{534}   \\
			\bottomrule 
			\multicolumn{12}{l}{\footnotesize{(\dag) The synthesis dataset is generated by combining ideal imaging conditions with only additive Gaussian white noise.}}\\
			\multicolumn{12}{l}{\footnotesize{(\ddag) The real-world dataset is collected by a near-Earth orbit space-based observation platform, and the imaging deviates from the ideal conditions \cite{11029687}.}}\\
			\multicolumn{12}{l}{\footnotesize{(*) The reported time refers to the average inference time per sequence.}}
		\end{tabular} \label{tab:SOTA2}
		 }
\end{table*}

\subsubsection{Quantitative Evaluation}
The results on the multi-frame infrared small target detection and space debris detection tasks are shown in Tab. \ref{tab:SOTA1} and \ref{tab:SOTA2}. Multi-frame methods outperform single-frame ones, and deep learning approaches consistently surpass model-driven methods. Infrared small target detection methods that rely on structural priors of the target perform poorly on real astronomical images. In contrast, our method achieves superior performance on both tasks.

We further quantify the spatial extent of targets in our datasets. In the NUDT-MIRSDT and NUDT-MIRSDT-HiNo datasets, the target region occupies an average relative area of $5.949\times10^{-4}$, with the largest target accounting for 
$2.1725\times10^{-3}$. These values are well within the typical definition of small targets. In the SSTD-R dataset, the targets are even more extreme: the median relative area is $2.3836\times10^{-5}$, and the maximum reaches only 
$8.1057\times10^{-5}$. Under these conditions, the stride-4 downsampling in LSE reduces the spatial resolution to one-quarter of the original, effectively suppressing most target responses before they enter the Transformer blocks, allowing the Transformer to focus primarily on background representation rather than target discrimination. Moreover, our concurrent work \cite{zhang2026effective} further demonstrates that large effective receptive fields exhibit an inherent low-frequency bias, such that even in the rare case where target information survives the downsampling, it is progressively attenuated during forward propagation.

Due to the extremely low target signal-to-noise ratio and the presence of strong noise interference in NUDT-MIRSDT and NUDT-MIRSDT-HiNo, several single-frame methods fail to converge. Although the DTUM method focuses on the spatiotemporal motion cues of the target, accurately extracting such cues from weak targets under heavy noise is highly challenging. The DeepPro method enhances target saliency via temporal slicing, but its lack of spatial information leads to degraded detection performance. In particular, leveraging both local neighborhood and non-local information proves effective in suppressing strong noise. In comparison, our method achieves the best performance, as it emphasizes background discrimination by treating the target as an anomaly. The low-rank nature of the background serves as a stable and reliable prior across diverse conditions. The ROC curves presented in Fig. \ref{fig:ROC} demonstrate the robustness of the proposed method.

Synthesis astronomical images are generated under idealized conditions, which implicitly introduce strong priors about the target. As a result, models that focus on target-specific information tend to perform well on simulated data but struggle on real astronomical images. This performance gap arises because the imaging process in real scenarios is influenced by the complex interplay between space-based observation platforms and solar radiation, which can lead to abnormal variations in stellar intensity. Moreover, real astronomical images are affected by complex noise patterns that go far beyond standard Gaussian white noise. In contrast, our method demonstrates stronger adaptability by identifying targets through the discrimination of stars. 

\subsubsection{Qualitative Evaluation}
\begin{figure*}[!t]
	\centering
	\includegraphics[width=1\linewidth]{IR_Result.pdf}
	\caption{Visual comparison on the NUDT-MIRSDT dataset ($SNR \le 3$). For better visualization,the target area is enlarged in the top-right corner and highlighted with a \textcolor{red}{red} circle. The false alarm area is marked with a \textcolor[rgb]{1.0, 0.84, 0.0}{yellow} circle.}
	\label{fig:IR_Result}
\end{figure*}
\begin{figure*}[!t]
	\centering
	\includegraphics[width=1\linewidth]{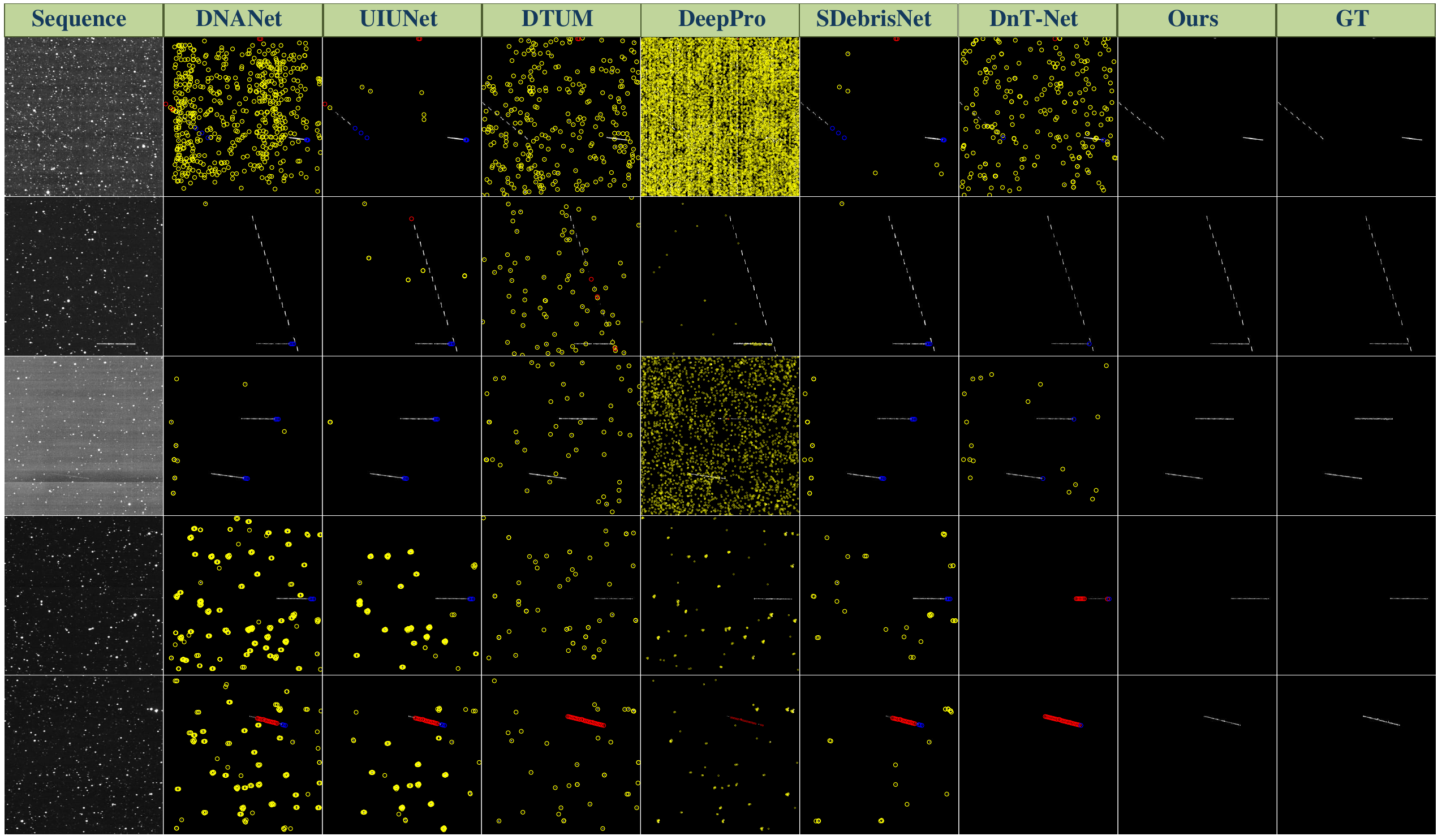}
	\caption{Visual comparisons on five sequences from real optical dataset are provided, where the raw images and detection results are overlaid for better illustration. Unlike simulated data, real-world imaging conditions are considerably more complex. The intensity of stars may fluctuate drastically due to variations in viewing angles and imaging geometry, and the image noise is far from ideal Gaussian white noise, often exhibiting structured or sensor-specific characteristics. \textcolor{blue}{Blue}, \textcolor[rgb]{1.0, 0.84, 0.0}{yellow} and \textcolor{red}{red} circles indicate true targets omitted by initialization, false alarms and missed detections, respectively.}
	\label{fig:RSO_Result}
\end{figure*}
Qualitative results are presented in Fig. \ref{fig:IR_Result} and Fig. \ref{fig:RSO_Result}, where it can be intuitively observed that our method exhibits strong adaptability to extremely weak infrared small targets in complex scenes. In particular, by jointly modeling both local and non-local spatiotemporal relationships, our approach achieves superior structural integrity in the detected targets. At this stage, the performance of deep learning-based methods \cite{Dai_2021_WACV} even falls behind that of model-driven approaches \cite{li2023sparse}. This is because model-driven methods typically employ low-rank and sparse decomposition to discriminate the background and thereby identify potential targets. However, since such methods only capture structural priors rather than semantic information, the resulting target segmentation tends to be suboptimal. Although DeepPro leverages temporal saliency via time-series profiling to enhance weak signal detection, its limited capacity to capture spatial information leads to inferior segmentation performance. For military early warning applications \cite{zhao2022single}, precise target localization heavily relies on structural cues such as shape, further emphasizing the importance of maintaining target integrity.

In the space debris detection task, a single frame often fails to reveal the debris target clearly due to its weak signal and transient appearance. Therefore, we aggregate the detection results over the entire sequence for visualization.
Due to interactions with solar radiation, stars may exhibit abnormal intensity fluctuations across frames, which often mislead contrast-based algorithms and result in a high number of false alarms. In contrast, our method focuses on the self-similarity among stellar patterns rather than on the individual appearance of stars, leading to greater robustness against such fluctuations.
For example, in the first row of the visualization, a debris target appears only in two frames before leaving the camera’s field of view. Algorithms that rely on motion cues fail to detect it due to the insufficient temporal support. However, our method treats the target as a spatial–temporal outlier and leverages the global sequence statistics to model stellar structures, allowing for reliable target discrimination.
Moreover, our approach does not rely on initialization or frame discarding strategies that are often employed to stabilize early-stage results. In contrast, methods \cite{cite-sdebrisnet,chen2024convolutional} specifically designed for space debris detection typically require initialization, leading to the omission of the first few frames. It is capable of detecting extremely weak targets whose spatial signatures are nearly indistinguishable. In such cases, motion-based cues become unreliable, while treating the target as an outlier from a static perspective provides a more robust and principled detection strategy.

\subsection{Cross-Domain Generalization Analysis}
\begin{figure*}[!t]
	\centering
	\includegraphics[width=1\linewidth]{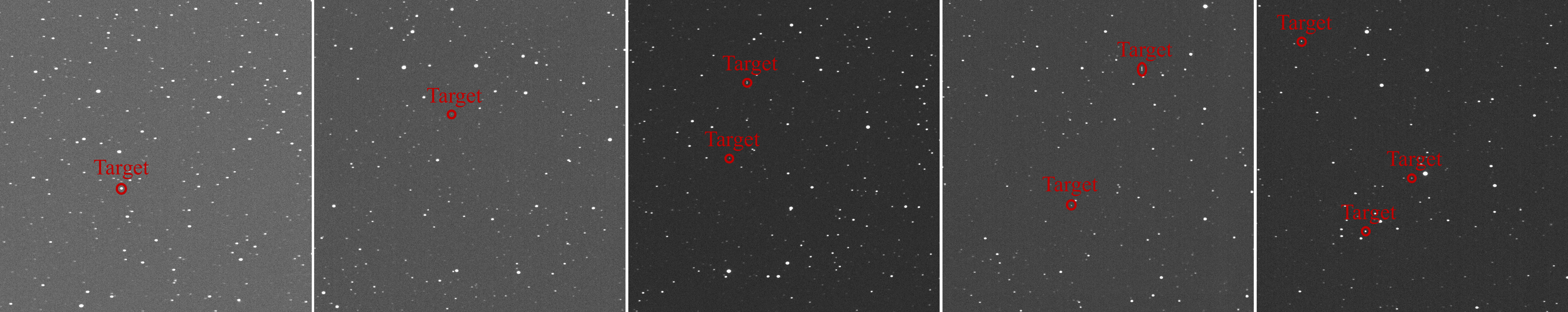}
	\caption{Representative sequence illustration for cross-dataset evaluation. To assess the performance across datasets, we employ \textbf{real star field images acquired by the ground-based observation platform of the Changchun Observatory, National Astronomical Observatories, Chinese Academy of Sciences.} The ground-truth annotations of targets are \textbf{derived from back-projections of catalogued objects}. As shown, targets and stars exhibit no distinguishable features in any single frame; they differ solely in their motion patterns.}
	\label{fig:RSO_New}
\end{figure*}
For cross-dataset generalization testing, we use real star-field data from the ground-based platform at the Changchun Observatory (NAOC), with representative examples in Fig.~\ref{fig:RSO_New}. Our model is trained on SSTD-R, which comprises data from a low-Earth-orbit space-based platform.
\begin{table}[]
	\centering
	\caption{Cross-Domain Generalization Analysis}
	\begin{tabular}{c|ccc}
		\toprule
		Methods & $\mathrm{R}^t$ ($\uparrow$)        & $\mathrm{FA}^t$ ($\downarrow$)   & $\mathrm{F1}^t$ ($\uparrow$)                                                      \\ \midrule
		 DeepPro & 90.00\% & $\approx$100.0\% &$\approx$0\% \\
		 TenRPCANet & 93.35\% &5.765\% &93.79\%  \\
		\bottomrule
	\end{tabular} \label{Tab:Cross-Domain}
\end{table}
The results are presented in Tab. \ref{Tab:Cross-Domain}. DeepPro achieves an F1‑score close to zero, which can be attributed to the following factors. (1) The degradation factors affecting space‑based and ground‑based imaging are inherently different, causing a substantial shift in the learned temporal profiles. (2) More critically, space debris does not emit energy on its own; its observability relies on reflected sunlight. Consequently, the received energy differs between space‑based and ground‑based platforms due to distinct attenuation mechanisms. (3) Unlike infrared small‑target imaging, where the target emits thermal radiation and thus exhibits inter‑frame consistency, space debris is subject to orbital mechanics and inevitably spins in orbit. Moreover, since most debris results from collisions, its shape and surface reflectance vary across different facets, leading to inconsistent target signatures across frames. (4) During imaging, neither space‑based nor ground‑based systems adopt a constant exposure time; as a result, the apparent scale of the same target varies from frame to frame, as its morphology is modulated by the exposure duration. In contrast, our method maintains high performance, because, as analyzed in Sec. \ref{Sec:Background}, we explicitly model the topological relationship consistency among stars, which is a physically guaranteed property in the real world, while treating the target as an outlier.
\subsection{Comparison and Discussion with Existing Space Debris Detection Methods}
\begin{figure}[!t]
	\centering
	\includegraphics[width=1\linewidth]{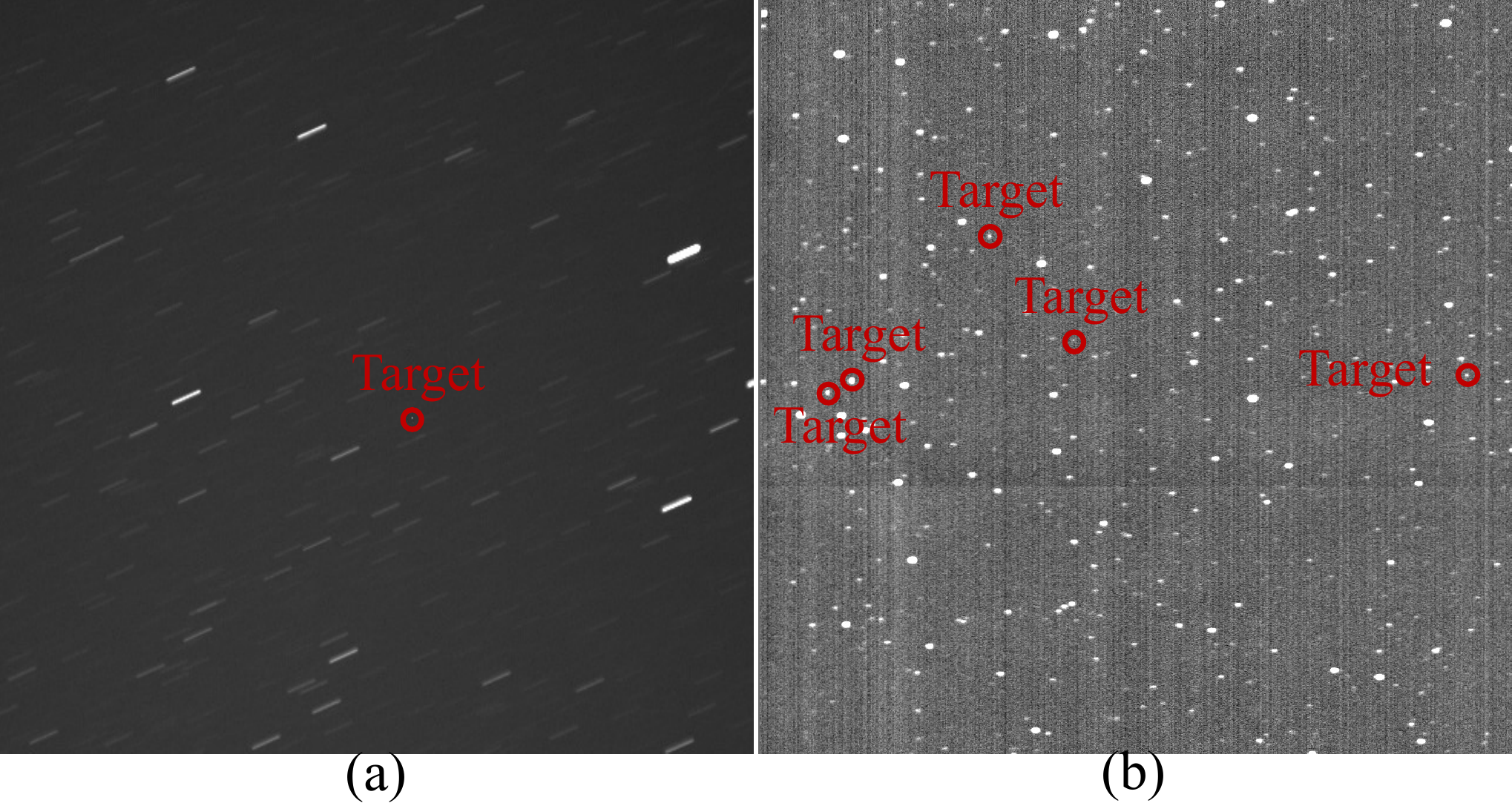}
	\caption{Comparison of existing space debris detection datasets and our representative images. (a) Existing data; (b) SSTD-R. Since space-based observation data are regulated and not publicly accessible, existing datasets are predominantly ground-based. In addition, large ground-based platforms are subject to restrictions by national observatories, so conventional research can only employ small-aperture cameras. To increase the limiting magnitude (a specific metric in astronomy), extremely long exposure times (several seconds to minutes per image) are required. Consequently, targets and stars exhibit significant differences within a single frame.}
	\label{fig:RSO_Com}
\end{figure}
\begin{table}[]
	\centering
	\caption{Comparison with Existing Space Debris Detection Methods}
	\begin{tabular}{c|ccc}
		\toprule
		Methods & $\mathrm{R}^t$ ($\uparrow$)        & $\mathrm{FA}^t$ ($\downarrow$)   & $\mathrm{F1}^t$ ($\uparrow$)                                                      \\ \midrule
		STDNet \cite{zhang2026stdnet} & 30.00\% & 91.93\% &12.72\% \\
		TenRPCANet & 99.53\% &1.000\% &99.00\%  \\
		\bottomrule
	\end{tabular} \label{Tab:Existing_Space_Debris_Detection}
\end{table}
As shown in Fig.~\ref{fig:RSO_Com}, \textbf{access to space debris astronomical images is severely restricted, especially for space‑based data under military control.} Most researchers therefore rely on ground‑based small‑aperture imaging, which requires long exposures (seconds to minutes) to detect faint targets. Under such long exposures, however, targets and stars become clearly distinguishable even in a single frame \cite{fu2024oriented}, so most existing detection methods are designed for single‑frame inputs and are thus inapplicable to our dataset \cite{fu2025federated}. Additionally, the three typical observation modes impose strict exposure constraints, further limiting practical deployment. In Tab.~\ref{Tab:Existing_Space_Debris_Detection}, we compare with the state‑of‑the‑art STDNet~\cite{zhang2026stdnet} on the space‑based SSTD‑R dataset. STDNet performs poorly because real space‑based data are far more complex than the ground‑based data it was trained on, and detection‑based supervision is sparser than segmentation, which makes convergence more difficult. In contrast, our segmentation‑based design is motivated by the reliability of space‑based photometry. Since the target is governed by orbital mechanics and spins in orbit, accurate intensity extraction is essential for characterizing its status, and segmentation provides exactly that capability.

\subsection{Discussion}
\noindent\textbf{On Novelty.} While it is widely acknowledged that target detection and background discrimination are two sides of the same coin, how to instantiate this principle in practice remains ambiguous. A critical yet often overlooked issue is that the proper formulation of low-rank and sparse decomposition is task-dependent. For instance, infrared small target detection conventionally employs patch-based tensor construction \cite{zhang2019infrared}, whereas our prior investigation \cite{wang2025anomalous} reveals that such patch-based modeling is inappropriate for space debris detection, where all point sources—including both stars and potential targets—must be extracted holistically. This fundamental discrepancy precludes a unified deep unfolding framework across tasks. Moreover, the annotation of all point sources is practically infeasible for deep learning due to the sheer density and scale of astronomical images. In contrast, our method circumvents these difficulties by demonstrating that self-attention naturally models low-rank structures and inherently inherits rank constraints from token representations, offering a more flexible and practical solution.

\noindent\textbf{On Generalizability.} We define generalizability along three dimensions. First, methods specifically designed for infrared small target detection perform poorly on space debris detection, whereas our approach achieves competitive performance on both tasks simultaneously. Second, many existing space debris detection algorithms are contingent upon restrictive assumptions—they either assume that stars appear as points and targets as streaks, or vice versa. In real-world wide-field imaging, however, both manifestations co-exist (as illustrated in Fig. \ref{fig:RSO_Result}), rendering such assumptions brittle. Third, our model, trained exclusively on space-based data, generalizes effectively to ground-based observations, demonstrating cross-platform transferability.

\noindent\textbf{On Paradigm Shift.} Although the coupled nature of target-background discrimination has long been recognized in traditional model-driven approaches, contemporary deep learning methods predominantly focus on learning target-specific features \cite{li2025probing,li2023direction,ying2025infrared}. This preoccupation with foreground characteristics, while neglecting the structural regularities of the background, represents a missed opportunity. Our work reintroduces the background discrimination perspective into the deep learning paradigm, marking a conceptual departure from prevailing trend.

\noindent\textbf{On Implicit Low-Rank Modeling.} It is essential to clarify that low-rankness is an intrinsic property of the data rather than an imposed constraint. While different scenarios may share the general property of low-rank backgrounds, explicit optimization—such as computing singular values or performing tensor decompositions—inevitably tailors the model to the specific statistical characteristics of the training data, thereby compromising generalization. This limitation is evident in classical tensor RPCA methods, where the low-rank and sparse components must be re-optimized from scratch for each new sequence, and has been further exposed in recent deep unfolding approaches that rely on explicit rank functionals \cite{yin2026dnn}. In stark contrast, our method enables the self-attention mechanism, under the structural regularization of LSE, to learn a discriminative low-rank representation that is not tied to any fixed rank functional. This implicit formulation preserves the flexibility to adapt to varying background statistics, underpinning the strong cross-task and cross-platform generalization we observe.

\noindent\textbf{ On Flexibility.} Our study further reveals that the self-attention mechanism offers a flexible framework for low-rank discrimination (as shown in Theorem \ref{thm:low_rank_attention}), a property not readily available in conventional tensor-based formulations. Specifically, different tokenization strategies can directly inject distinct low-rank constraints into the attention process, enabling the model to adapt to diverse observation geometries and data characteristics. This flexibility is particularly significant when contrasted with deep unfolding approaches, which are inherently constrained by the fixed tensor order of their constructed observations. In the model-driven era, different tasks necessitate different tensor organizations, and these incompatible formulations cannot be unified within a single optimization objective. Our framework resolves this long-standing limitation by decoupling the tokenization strategy from the subsequent low-rank modeling. The self-attention mechanism remains agnostic to the specific tokenization employed, while the tokenization itself provides the task-appropriate structural prior. This modularity allows the same architectural backbone to accommodate disparate data modalities and observation conditions without sacrificing theoretical coherence or task-specific performance.

\subsection{Ablation Study}
\begin{table*}[]
	\centering
	\caption{Quantitative ablation study assessing the impacts of core components LSE and PFR. The metrics considered include $F_a$ ($10^{-5}$).}
	\begin{tabular}{c|cc|ccc|cccc|ccc}
		\toprule
		\multirow{2}{*}{Strategy} & \multicolumn{2}{c|}{Module} & \multicolumn{3}{c|}{NUDT-MIRSDT} & \multicolumn{4}{c|}{Real Optical Set} & \multicolumn{3}{c}{Scale\dag} \\
		& LSE          & PFR          & $P_d$ ($\uparrow$)        & $F_a$ ($\downarrow$)       & AUC ($\uparrow$)       & $\text{R}^t$ ($\uparrow$)     & $\text{FA}^t$ ($\downarrow$)      & $\text{F1}^t$ ($\uparrow$)     & IoU ($\uparrow$)     & Param (M)  & GFLOPs & FPS \\ \midrule
		(a) & \XSolidBrush & \XSolidBrush & 30.14\% & 1.02 & 0.8863 & \multicolumn{4}{c|}{Not Converging} & 1.70 & 2.92 & 603.77 \\
		(b) & \CheckmarkBold & \XSolidBrush & 85.43\% & 2.66 & 0.9879 &\multicolumn{4}{c|}{Not Converging} & 1.76 & 16.93 & 187.36 \\ 
		(c) &\Checkmark\kern-1.2ex\raisebox{1ex}{\rotatebox[origin=c]{125}{\textbf{--}}}\ddag  & \CheckmarkBold & 97.98\% & 8.82 & 0.9907 & 99.53\% & 1.8\% & 98.80\% & 77.52\% & 1.77 & 18.83 & 178.40 \\
		(d)  & \CheckmarkBold &\Checkmark\kern-1.2ex\raisebox{1ex}{\rotatebox[origin=c]{125}{\textbf{--}}}¶ & 96.39\% & 7.03 & 0.9921 & 95.65\% & 20.89\% & 86.05\% & 52.40\% & 1.78 & 18.73 & 177.56 \\
		(e)  & \CheckmarkBold &\Checkmark\kern-1.2ex\raisebox{1ex}{\rotatebox[origin=c]{125}{\textbf{--}}}\# & 99.65\% & 8.9 & 0.9973 & 98.06\% & 8.58\% & 94.35\% & 78.30\% & 1.78 & 19.14 & 177.12 \\
		\rowcolor{gray!20}
		(f) & \CheckmarkBold & \CheckmarkBold & 99.33\% & 0.36 & 0.9978 & 99.53\% & 1.0\% & 99.00\% & 80.78\% & 1.78 & 20.43 & 176.24 \\
		(g) & \CheckmarkBold \S & \CheckmarkBold & 99.25\% & 27.96 & 0.9927 & 86.79\% & 11.18\% & 86.73\% & 47.34\% & 1.78 & 20.43 & 176.24 \\
		(h) & \CheckmarkBold & \CheckmarkBold* & 99.65\% & 17.14 & 0.9992 & 63.33\% & 49.03\% & 49.34\% & 9.39\% & 1.78 & 20.43 & 176.24 \\
		(i) & \CheckmarkBold \S & \CheckmarkBold* & 99.48\% & 9.14 & 0.9942 & 92.76\% & 11.37\% & 90.35\% & 42.40\% & 1.78 & 20.43 & 176.24 \\
		\bottomrule
		\multicolumn{12}{l}{\footnotesize{(\dag) The metrics are recorded on the multi-frame infrared small target detection task.}} \\
		\multicolumn{12}{l}{\footnotesize{(\ddag) Since the PFR module depends on the output of the LSE module, only the 2D patch branch of LSE is preserved.}}\\
		\multicolumn{12}{l}{\footnotesize{(¶) Removing the 3$\times$3 convolution in the PFR module suppresses its ability to perceive structural sparsity.}}\\
		\multicolumn{12}{l}{\footnotesize{(\#) Retaining only the final 3$\times$3 convolution in the PFR module suppresses its ability to perceive structural sparsity.}}\\
		\multicolumn{12}{l}{\footnotesize{(\S) Nonlinear activation functions (ReLU) are incorporated into the proposed LSE module.}}\\
		\multicolumn{12}{l}{\footnotesize{(*) Nonlinear activation functions (ReLU) are incorporated into the proposed PFR module.}}
	\end{tabular} \label{Tab:ALL}
\end{table*}

\begin{table}[]
	\centering
	\caption{Ablation Study on the Branches of the LSE Module.}
	\begin{tabular}{cccc|ccc|c}
		\toprule
		\multicolumn{4}{c|}{Branch} & \multicolumn{3}{c|}{NUDT-MRISTD} & \multirow{2}{*}{Param} \\
		3$\times$3     & 5$\times$5     & 7$\times$7    & 9$\times$9    & $P_d$ ($\uparrow$)        & $F_a$ ($\downarrow$)       & AUC ($\uparrow$)       &                            \\\midrule
		\CheckmarkBold & \XSolidBrush & \XSolidBrush & \XSolidBrush & 90.74\% & 15.42 & 0.9901 & 1.70 \\
		\CheckmarkBold & \CheckmarkBold & \XSolidBrush & \XSolidBrush & 91.52\% & 4.635 & 0.9913 & 1.73 \\ \rowcolor{gray!20}
		\CheckmarkBold & \CheckmarkBold & \CheckmarkBold & \XSolidBrush & 99.33\% & 0.36 & 0.9978 & 1.78 \\
		\CheckmarkBold & \CheckmarkBold & \CheckmarkBold & \CheckmarkBold & 99.28\% & 5.731
		 & 0.9972 & 2.63 \\
		\CheckmarkBold & \CheckmarkBold & \XSolidBrush & \CheckmarkBold & 99.45\% & 34.01 & 0.9981 & 2.11 \\
		\CheckmarkBold & \XSolidBrush & \CheckmarkBold & \CheckmarkBold & 88.37\% & 6.545 & 0.9886 & 2.59 \\
		\XSolidBrush & \CheckmarkBold & \CheckmarkBold & \CheckmarkBold & 98.12\% & 6.132 & 0.9898 & 2.62 \\
		 \bottomrule
	\end{tabular} \label{Tab:Linear}
\end{table}

\begin{table}[]
	\centering
	\caption{Ablation Study on Spatiotemporal Window Size}
	\begin{tabular}{c|c|ccc|c}
		\toprule
		\multirow{2}{*}{In $T$-frames} & \multirow{2}{*}{Window-Size} & \multicolumn{3}{c|}{NUDT-MRISTD} & \multirow{2}{*}{FPS} \\
		&                              & $P_d$ ($\uparrow$)        & $F_a$ ($\downarrow$)       & AUC ($\uparrow$)       &                                                   \\ \midrule
		2 & 2$\times$7$\times$7 & 87.01\% &15.44 &0.9857 &228.12 \\
		4 & 2$\times$7$\times$7 & 89.33\% &11.12 &0.9888 &210.74 \\
		4 & 4$\times$7$\times$7 & 98.21\% &1.88 &0.9964 &206.61 \\
		8 & 2$\times$7$\times$7 & 92.11\% &8.61 &0.9867 &180.34 \\
		8 & 4$\times$7$\times$7 & 98.88\% &1.956 &0.9911 &178.60 \\\rowcolor{gray!20}
		8 & 8$\times$7$\times$7 & 99.33\% & 0.36 & 0.9978 &176.24 \\
		\bottomrule
	\end{tabular} \label{Tab:LSE_B}
\end{table}

\begin{table}[]
	\centering
	\caption{Ablation Study on Model Hyperparameters}
	\begin{tabular}{c|c|c|ccc}
		\toprule
		\multirow{2}{*}{Channel   Size} & \multirow{2}{*}{Param} & \multirow{2}{*}{GFLOPs}  & \multicolumn{3}{c}{NUDT-MRISTD} \\
		&                        &                                              & $P_d$ ($\uparrow$)        & $F_a$ ($\downarrow$)       & AUC ($\uparrow$)       \\\midrule
		12 &0.46 &5.247 &98.82\% &3.72 &0.9954 \\\rowcolor{gray!20}
		24 &1.78 &20.43 &99.33\% & 0.36 & 0.9978 \\
		48 &6.94 &79.68 &98.89\% & 0.26 &0.9969 \\
		 \bottomrule
	\end{tabular} \label{Tab:Size}
\end{table}
\subsubsection{Ablation Study on Key Components} Our model’s performance heavily relies on two key components: the Locally Subspace Embedding (LSE) and the Progressive Feature Refinement (PFR) module. Both work synergistically with the encoder-decoder architecture to realize the tensor low-rank and sparse decomposition. Notably, the PFR module cannot function effectively in isolation without LSE, indicating a strong interdependency. Ablation results are presented in Tab. \ref{Tab:ALL}. Without the LSE and PFR modules, the model fails to converge on the space debris detection task. Comparing strategies (c) and (f) reveals that the 3D patch construction branch within the LSE module effectively suppresses false alarms. This improvement stems from the introduction of a fourth-order tensor low-rank prior enabled by the 3D patch grouping, which captures the intrinsic spatiotemporal structure of the video sequence.

Comparing strategies (d), (e) and (f) reveals that removing the structural sparsity modeling capability from the PFR module results in a significant performance drop in space debris detection. This is primarily because, compared to infrared small target detection, astronomical imagery contains more complex interference sources and much stronger \cite{11029687}, non-Gaussian noise, making the task considerably more challenging.

In our original design, both the LSE and PFR modules are entirely linear. This design choice aims to prevent overfitting to shallow, specific patterns, which could degrade model performance on unseen scenarios. Ablation experiments comparing strategies (g) through (i) confirm this hypothesis, demonstrating that introducing non-linearities in these modules leads to performance drops, thereby validating the effectiveness of maintaining their linearity.

\subsubsection{Ablation Study on the Multi-Branch Design of the LSE} Inspired by the superior performance of tensor low-rank and sparse decomposition following multilinear tensor construction in model-driven approaches \cite{4032832,6482624}, we incorporate a multi-branch design within the LSE module. While the previous subsection presented an ablation study on the effectiveness of linear constructions, this section focuses on evaluating the contribution of the multi-branch architecture. The ablation results are summarized in Tab. \ref{Tab:Linear}.

Since different types of false alarms often require distinct contextual information to be effectively suppressed \cite{li2024lsknet}, adopting a multi-branch architecture is well-suited for this task. In addition, as the spatial scale increases, the optimization becomes more challenging due to the larger search space and increased complexity. On the other hand, neglecting small local windows may lead to the loss of fine-grained details that are crucial for accurately identifying small targets.

\subsubsection{Ablation Study on Hyperparameters}
The hyperparameter analysis, as shown in Tab. \ref{Tab:LSE_B} and \ref{Tab:Size}, indicates that increasing the number of input frames and enlarging the temporal window both lead to improved detection performance. This is because a longer temporal dimension provides richer contextual information, which helps to better distinguish the background and reduces the influence of occasional anomalies.
However, in many real-world applications \cite{gao2013infrared}, it is often challenging to obtain long video sequences due to operational constraints. For example, airborne infrared observation systems typically cannot maintain long-term staring at a fixed region \cite{zhao2022single}; instead, they rely on wide-area search enabled by the motion of electro-optical (EO) pods. To accommodate these practical requirements, we set the temporal window to 8 frames in our experiments.

\subsubsection{Discussion on Memory Consumption}
We next examine inference GPU memory for the optimal window size $8\times7\times7$ (8-frame clips). \textbf{For high-resolution ($1024\times1024$) inputs in space debris detection, the feature map $1\times1\times8\times1024\times1024$ consumes only 4.8 GB}, mainly due to LSE downsampling to $256\times256$ and subsequent 3D Swin blocks with $8\times7\times7$ windows and hierarchical downsampling. Beyond image-level detection, space debris tasks require precise localization, orbit determination, and cataloging, often with multi-sensor coordination; however, inference can be run on ground servers with ample resources.

\subsection{Limitations}
\begin{figure}[!t]
	\centering
	\subfloat[]{
		\includegraphics[width=\linewidth]{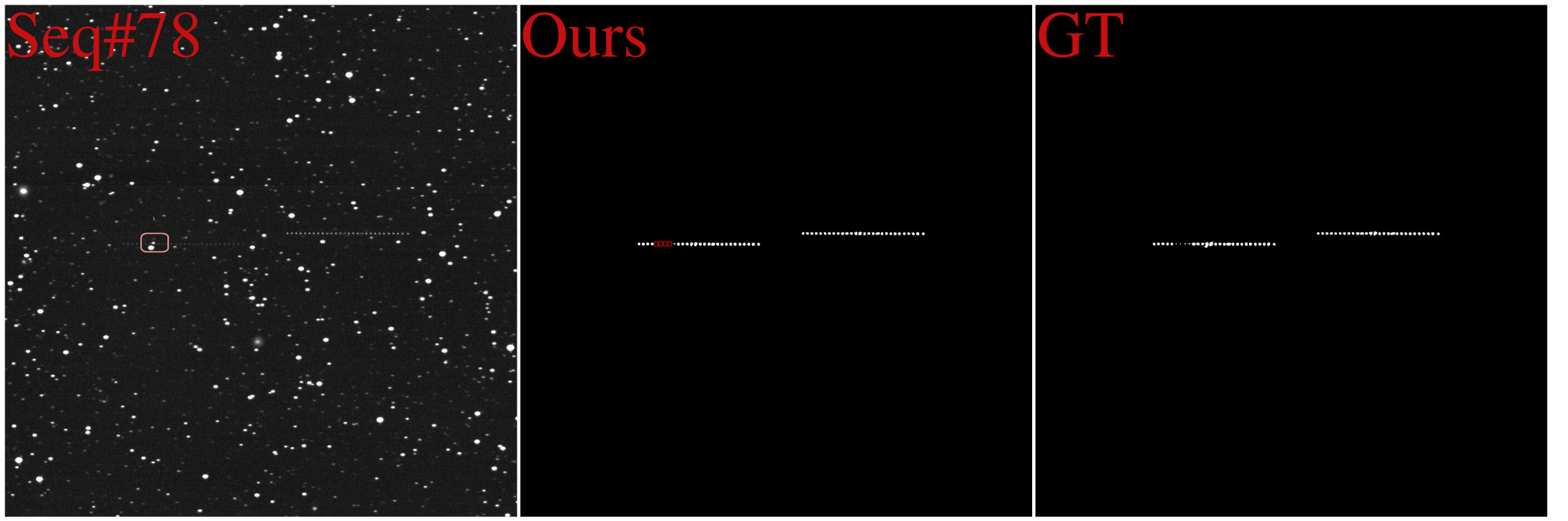}
		\label{fig:Failed}
	}
	\hfill
	\vspace{-7pt} 
	\subfloat[]{
		\includegraphics[width=\linewidth]{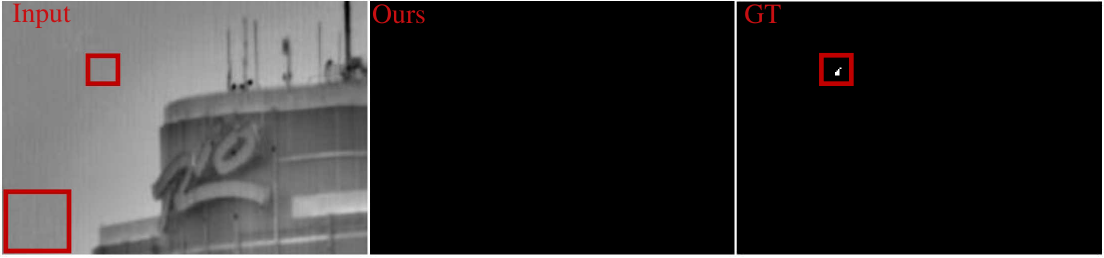}
		\label{fig:Failed2}
	}
	\caption{Typical failure cases. (a) The detection fails when the target overlaps with a star. (b) The detection fails when the target signal is very weak.}
	\label{fig:main}
\end{figure}
Typical failure cases are shown in Fig. \ref{fig:Failed} and \ref{fig:Failed2}. The causes of these failures stem from our method treating the input $T$-frame sequence equally. Specifically, focusing on short-term information helps distinguish targets obscured by stars, while focusing on long-term information enables the transformation of weak signals into prominent signals in the temporal profile. In the future, we will explore the design of spatiotemporal cooperative self-attention to adaptively leverage both long-term and short-term information.

\section{Conclusion}\label{Section:Conclusion}
In this paper, we propose a novel deep learning paradigm for small moving target detection, which leverages the low-rank property of the background while relaxing conventional assumptions on foreground sparsity and motion cues. We conduct a physically grounded analysis of the characteristics of background, target, and noise. Based on these insights, we design TenRPCANet, an end-to-end architecture that implicitly performs low-rank and sparse decomposition. Extensive experiments on multiple public datasets in two representative tasks validate the effectiveness of the proposed method. This work provides a solid foundation for the design of future deep learning-based small moving target detection algorithms.

For tasks where target morphology is of secondary importance, such as anti-UAV surveillance, detector-based formulations may be preferable as they reduce annotation burden. However, for video-based infrared small target detection, two fundamental challenges remain: (1) insufficient single-frame representation \cite{zhang2025you}, simply introducing high-resolution features often leads to shortcut learning rather than genuine discrimination; and (2) insufficient motion representation \cite{zhang2026decoupled}, implicitly learning the coupled motion of the camera, target, and scene tends to be dominated by background statistics during feature aggregation, necessitating explicit motion disentanglement. 
\bibliographystyle{IEEEtran}

\vspace{-45pt}
\begin{IEEEbiography}  [{\includegraphics[width=0.9in,height=1.125in,clip]{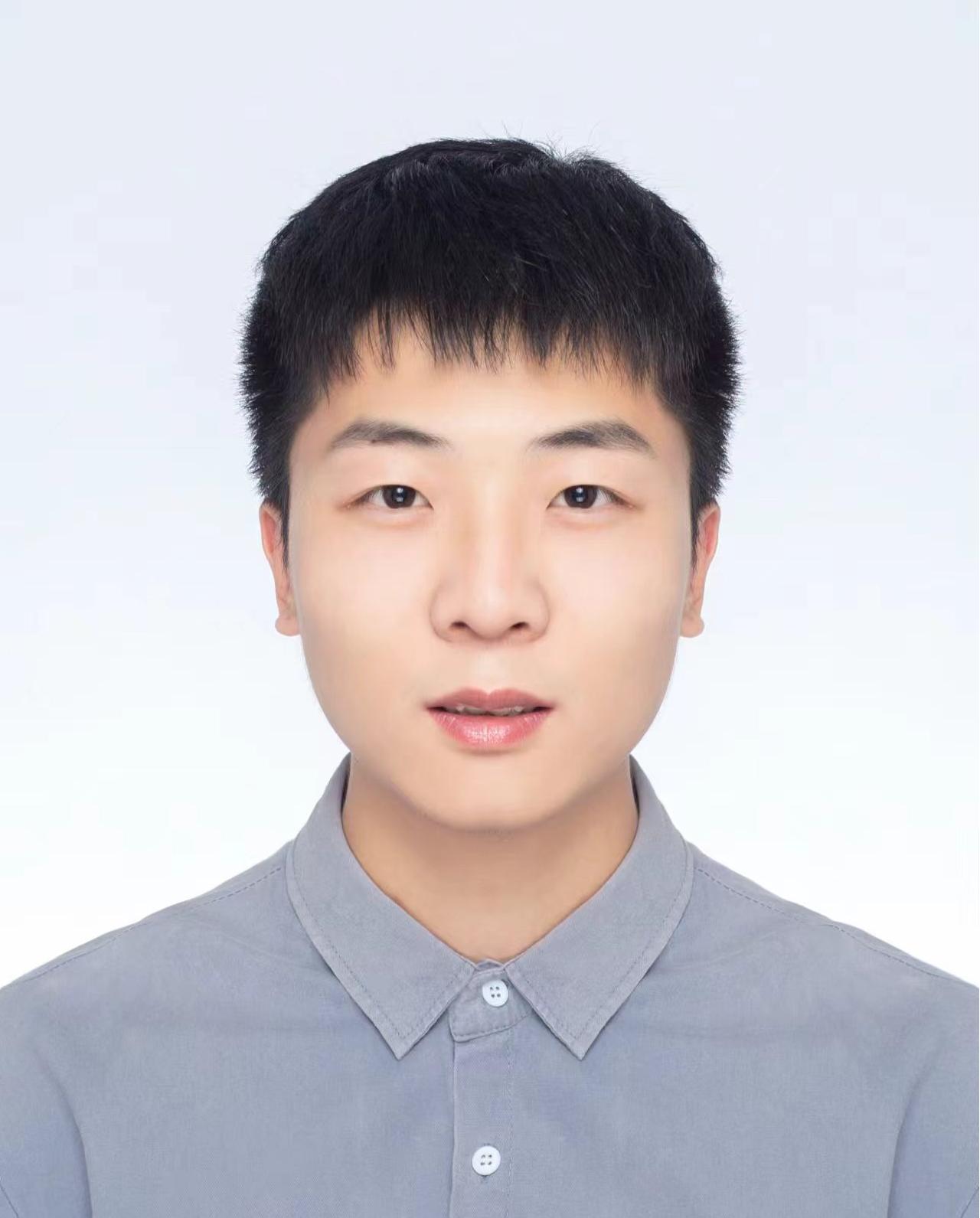}}] 
	{Guoyi Zhang} received the B.E. degree from Hefei University of Technology, Hefei, China, in 2023. He is currently pursuing the M.E. degree with the School of Aeronautics and Astronautics, Sun Yat-sen University, Shenzhen, China.\\
	His current research interests include representation learning and infrared small target detection. 
\end{IEEEbiography}
\vspace{-48pt}
\begin{IEEEbiography}  [{\includegraphics[width=1in,height=1.25in,clip,keepaspectratio]{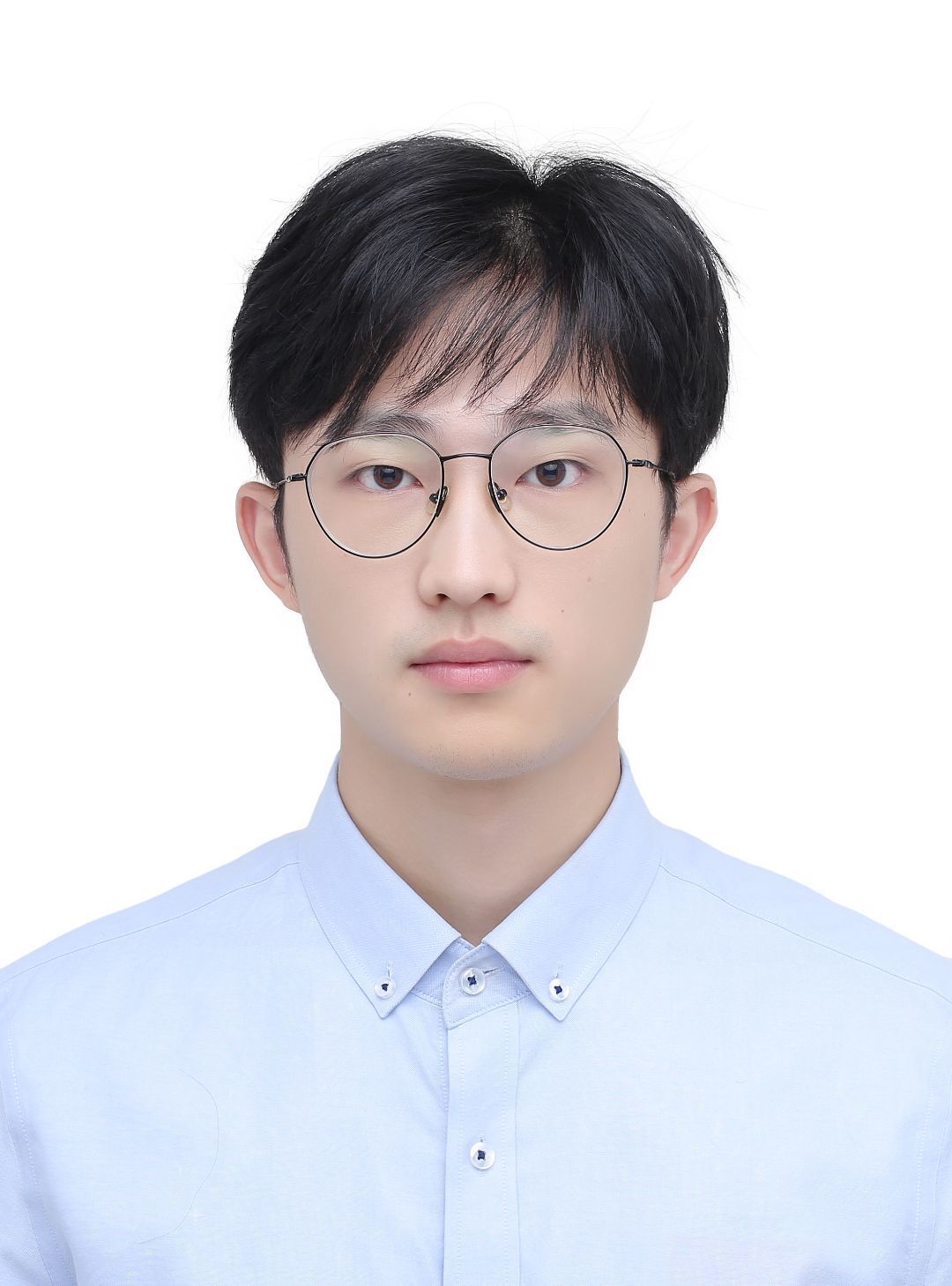}}] 
	{Han Wang} eceived the B.E. degree in mechanical engineering from Nanjing Agricultural University, Nanjing, China, in 2020, and the M.E. degree in mechanical engineering from China Agriculture University, Beijing, China, in 2022. He is currently working toward the Ph.D. degree with the School of Aeronautics and Astronautics, Sun Yat-sen University, Shenzhen, China.\\
	His research interests include astronomical image processing and space debris detection.
\end{IEEEbiography} 
\vspace{-48pt}
\begin{IEEEbiography}  [{\includegraphics[width=0.9in,height=1.125in,clip,keepaspectratio]{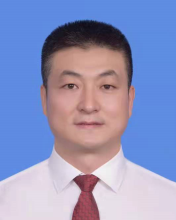}}] 
	{Xiaohu Zhang} received the Ph.D. degree in aeronautical and astronautical science and technology from National University of Defense Technology, Changsha, China, in 2006.\\
	He is currently a Professor with the School of Aeronautics and Astronautics, Sun Yat-sen University, Shenzhen, China. His research interests include aircraft visual perception, computer vision, and photogrammetry. 
\end{IEEEbiography} 
\newpage

\vfill

\end{document}